%% file: wsd-article.tex
\documentclass[a4paper,fleqn]{cas-sc}

\usepackage[numbers]{natbib}

\usepackage{ntheorem}

\usepackage{amsmath}
\usepackage{subcaption}

\graphicspath{{figs/pdf/}{figs/jpeg/}{figs/png/}{figs/svg/}{figs/bio/}}
\DeclareGraphicsExtensions{.pdf,.jpeg,.png,.svg}

\usepackage[export]{adjustbox}

\usepackage[justification=centering]{caption}

\usepackage{array,etoolbox}
\preto\tabular{\setcounter{magicrownumbers}{0}}
\newcounter{magicrownumbers}

\usepackage[acronym]{glossaries}
\include{abrev}

\usepackage{multirow}

\usepackage[linesnumbered,ruled,vlined]{algorithm2e}

\SetCommentSty{mycommfont}

\begin{document}
\let\WriteBookmarks\relax
\def\floatpagepagefraction{1}
\def\textpagefraction{.001}
\shorttitle{A Novel WSD Approach Using WordNet KG}
\shortauthors{M AlMousa et~al.}

\title[mode = title]{A Novel Word Sense Disambiguation Approach Using WordNet Knowledge Graph}

\author[1]{Mohannad AlMousa}[orcid=0000-0003-0087-1941]
\cormark[1]
\ead{malmous@lakeheadu.ca}
\credit{Conceptualization of this study, Methodology, Software, Investigation, Data Curation, Writing - Original Draft, Visualization}

\address[1]{Department of Software Engineering, Lakehead University, Thunder Bay,ON, P7B 5E1, Canada.}

\author[1]{Rachid Benlamri}[]
\ead{rbenlamr@lakeheadu.ca}
\ead[URL]{https://flash.lakeheadu.ca/~rbenlamr/}
\credit{Supervision, Validation, Writing - Review & Editing}

\author[2]{Richard Khoury}[]
\ead{richard.khoury@ift.ulaval.ca}
\ead[URL]{http://www2.ift.ulaval.ca/~rikho/}

\credit{Supervision, Validation, Writing - Review & Editing}

\address[2]{Department of Computer Science and Software Engineering, Université Laval, Québec, QC G1V 0A6, Canada.}

\cortext[cor1]{Corresponding author}

\begin{abstract}
Various applications in computational linguistics and artificial intelligence rely on high-performing word sense disambiguation techniques to solve challenging tasks such as information retrieval, machine translation, question answering, and document clustering.
While text comprehension is intuitive for humans, machines face tremendous challenges in processing and interpreting a human's natural language. 
This paper presents a novel knowledge-based word sense disambiguation algorithm, namely Sequential Contextual Similarity Matrix Multiplication (SCSMM). The SCSMM algorithm combines semantic similarity, heuristic knowledge, and document context to respectively exploit the merits of local context between consecutive terms, human knowledge about terms, and a document's main topic in disambiguating terms.
Unlike other algorithms, the SCSMM algorithm guarantees the capture of the maximum sentence context while maintaining the terms' order within the sentence. 
The proposed algorithm outperformed all other algorithms when disambiguating nouns on the combined gold standard datasets, while demonstrating comparable results to current state-of-the-art word sense disambiguation systems when dealing with each dataset separately.
Furthermore, the paper discusses the impact of granularity level, ambiguity rate, sentence size, and part of speech distribution on the performance of the proposed algorithm.
\end{abstract}

\begin{keywords}
\texttt{Semantic Word Sense Disambiguation}\sep \texttt{Knowledge-based}\sep \texttt{knowledge graph}\sep \texttt{WordNet}.
\end{keywords}

\maketitle

\section{Introduction}\label{sec:introduction}
Many \gls{nlp} applications rely on \gls{wsd}, either directly or indirectly. The list includes, but is not limited to \gls{mt}, \gls{ir}, \gls{qa}, \gls{ner}, and text summarization. \gls{wsd} is considered one of the oldest tasks of computational linguistics dating back to the 1940s. It started as a distinct task when machine translation was first developed. The first challenge that triggered \gls{wsd} task is \gls{mt} in the 1940s. Since then, researchers have been developing models and algorithms to improve the accuracy of this task using various approaches; supervised, semi-supervised, and knowledge-based systems. 
\gls{wsd} is an essential task in many other applications, such as \gls{ir}, information extraction, knowledge acquisition, and \gls{nlp}. With the introduction of supervised machine learning in the 1990s,  various supervised approaches attempted to solve the \gls{wsd} task. More recent studies are exploring semi-supervised and unsupervised approaches using knowledge base in the form of graph systems such as WordNet\footnote{\url{https://wordnet.princeton.edu/}} and BabelNet\footnote{\url{https://babelnet.org/}}.

\begin{figure}[!t]
	\centering
	\includegraphics[trim=20 190 40 200, clip, width=.7\columnwidth]{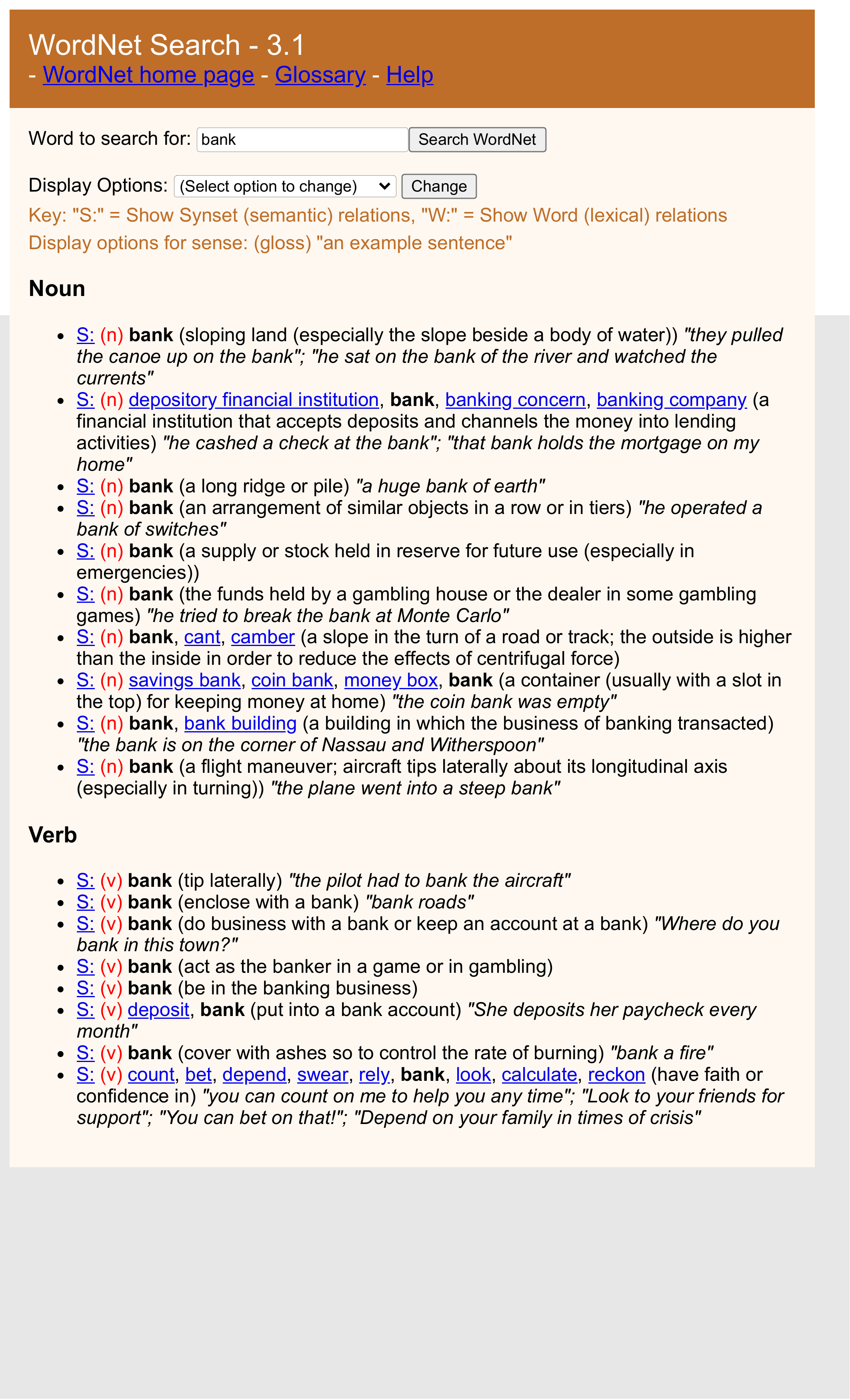}
	\caption{Senses for the term bank}
	\label{fig:BankSenses}
\end{figure}

Human beings can usually detect the appropriate sense unconsciously, whereas programming a machine to perform such a function is challenging. Within the \gls{nlp} domain, \gls{wsd} is the task to determine the appropriate meaning (sense) of words given a surrounding context. \gls{wsd} is considered a classification task, where the system's main task is to classify a specific word to one of its senses as defined by a lexical dictionary. One typical example is the word \textit{`bank'}, which has eighteen different senses defined in WordNet\footnote{\url{http://wordnetweb.princeton.edu/perl/webwn?s=bank}} lexical database, namely ten as a noun, and the rest as a verb, as shown in Fig. \ref{fig:BankSenses}.

\gls{wsd} systems are divided into four main categories based on their approach: supervised, semi-supervised, unsupervised, and knowledge-based. 
Supervised systems require a large sense-annotated training dataset. 
Semi-supervised systems employ a bootstrapping process with a small seed of a sense-annotated training dataset and a large corpus of un-annotated senses.
Unsupervised approaches use context clustering \cite{ji2010one}, word clustering \cite{pantel2002discovering}, or other graph-based algorithms such as the \textit{PageRank} algorithm \cite{agirre2009personalizing}.
Finally, knowledge-based approaches rely on the structure and features of a \gls{kg}, such as taxonomic relations, non-taxonomic relations, concept's \gls{ic}, and paths. 

Among all four \gls{wsd} categories, supervised and knowledge-based are the most promising approaches \cite{jurafsky2018ChC}. However, supervised approaches require a large annotated dataset, which is challenging to produce. Due to the limited number of sense-annotated datasets, these systems face challenges to excel and demonstrate a noticeable improvement over other systems. Moreover, for the most part, supervised systems require training dataset, in addition to being computationally expensive and time-consuming. Finally, most \gls{wsd} supervised systems are unable to intuitively explain their results since they usually use a training function that leads to a calculated decision-making process.

knowledge-based systems however, do not require a training dataset because they rely on a massive dictionary or \gls{kg}. Moreover, knowledge-based systems can easily explain their results since they normally follow an intuitive process. With the advancement of \gls{lod} and domain-specific \gls{kg}s, these systems have a higher potential to outperform other approaches due to the advantage of broader \gls{kg} coverage \cite{navigli2009word}, but achieving this requires a semantically rich \gls{kg} and a comprehensive semantic similarity measure.
The later is used to perform a \gls{wsd} task by assigning a weight to each sense of the ambiguous word based on its semantic similarity with other terms within the sentence, document, or both. The sense with the highest weight is selected as the correct sense.

In addition to the semantic similarity measure, word sense heuristics and document context are two important ingredients that have also been used in the literature for disambiguating words \cite{navigli2009word,raganato2017word,chaplot2018knowledge}. The word sense heuristic is expressed by the frequency distribution of the word's senses based on their usage in the training dataset (i.e., SemCor and OMSTI). The document context provides an ambiguous word with a global context that enables the selection of the appropriate sense. In this paper, we investigated the use of semantic similarity, word sense heuristics, and document context, to develop a novel knowledge-based \gls{wsd} algorithm, namely \gls{scsmm}. The proposed algorithm follows the disambiguation process of the human brain by exploiting the local context within the sentence, prior knowledge of the term's usage, and the global context of the document, which are represented by the semantic similarity between terms, terms frequency heuristics, and the document context, respectively. 

The rest of the paper is organized as follows: Section \ref{sec:Related Work} describes in detail the related work, motivations behind this study and contributions. Section \ref{sec:Proposed Method} introduces the proposed method for the \gls{wsd} system. In Section \ref{sec:Evaluation and Experimental Results}, we describe the experimental environment and discuss experimental results. Finally, conclusions are drawn and future research work is suggested in Section \ref{sec:Conclusion}.

\section{Related Work}\label{sec:Related Work}
The main objective of \gls{wsd} is to classify a word into its correct sense given a context. This task has been investigated within the computational linguistics field since the 1940s, and since then, many algorithms and techniques have been developed. 
\gls{wsd} is a challenging task for several reasons, one of which is related to the discrepancies of senses choices between dictionaries. One dictionary might provide more senses for a word than another. To overcome such a challenge, many researchers relied on a single comprehensive machine-readable lexical dictionary such as WordNet\footnote{\url{https://wordnet.princeton.edu/}}, Wikipedia\footnote{\url{https://www.wikipedia.org/}}, and BabelNet\footnote{\url{https://babelnet.org/}}. 

Another difficulty is derived from the evaluated test datasets and the inter-annotator agreement. The datasets to evaluate any system must be judged and annotated by humans because human judgment is considered a gold standard. Compiling test datasets is not an easy task, as it is difficult for humans to remember or know all senses for all words, including their precise meanings and differences from other senses. The gold standard datasets are usually measured by the inter-annotator agreement. Based on \cite{chklovski2003exploiting,palmer2007making,snyder2004english,navigli2009word,lacerra2020csi} the inter-annotator agreement using WordNet ranges between 67\% and 80\% on fine-grained inventory. Such a low range of inter-annotator agreement encouraged the research community to develop and further investigate coarse-grained databases. In fact, some of the coarse-grained inventory has achieved up to 90\% inter-annotator agreement \cite{gale1992estimating,navigli2009word,lacerra2020csi}. Nonetheless, significant effort has been made to compile high-quality datasets that are considered the primary gold standard for \gls{wsd} systems (i.e., SensEval2, SensEval3, SemEval 2007, SemEval 2013, and SemEval 2015). These datasets are further discussed in Section \ref{subsubsec:Evaluation Datasets (Gold Standard}.

A vast number of research approaches, techniques and models have attempted to solve the \gls{wsd} challenge as a standalone task or as part of a larger \gls{nlp} application \cite{liu2001disambiguating,navigli2009word,borah2014approaches,pal2015word,giyanani2013survey,sarmah2016survey,aliwy2019Word}. Either way, these approaches are grouped into four conventional categories. Supervised approaches require the use of a training dataset (i.e., a sense-annotated corpus). However, these corpora are hard to produce due to the complexity of identifying the best combination of words' senses based on their definitions from WordNet. 
To our knowledge, there are currently two such datasets available: SemCor \cite{miller1994SemCor} and \gls{omsti} \cite{taghipour2015OMSTI}, which will be discussed in Section \ref{subsubsec:Training Datasets}. 
Supervised approaches also require a \gls{ml} technique that will, through training, create a feature vector for each ambiguous word, train a classifier to appropriately assign the correct sense class to an ambiguous target word, and finally, test it using a dataset to evaluate the model \cite{vial2019sense,pasini2020train}.
Early development of supervised \gls{wsd} approaches include rule-based, probabilistic, or statistical models. Many comprehensive surveys have covered the mathematical details of each model in \cite{navigli2009word,giyanani2013survey,borah2014approaches,pal2015word,sarmah2016survey,aliwy2019Word}.  

Semi-supervised approaches take a middle ground strategy by using a secondary small sense-annotated corpus as seed data, then applying a bootstrapping process such as the one presented in \cite{mihalcea2004senselearner}. The bootstrapping technique requires only a small amount of tagged data that acts as seed data. This data then undergoes a supervised method to train an initial classifier, which is, in return, used on another untagged portion of the corpus to generate a larger training dataset. Only high-confidence classifications are considered as candidates for the final training dataset. Those same steps are then repeated in numerous iterations, and the training portion successively increases until the entire corpus is trained, or a maximum number of iterations caps the process. The main advantage of the bootstrapping approach is that it requires a small seed dataset to begin the training process. The seed data could be manually-annotated or generated by a small number of surefire decision rules. 	

Unlike the previous two categories, unsupervised approaches do not require prior knowledge of the text; hence, no manual sense-annotated corpus is required. Nonetheless, most techniques in this category still require a training corpus for an unsupervised training task. Algorithms from this group have been further categorized into three groups: context clustering, co-occurrence graphs, and word clustering. 
		
Finally, knowledge-based approaches do not require an intensive training process. However, they disambiguate words in context by exploiting large scale knowledge resources (i.e., dictionaries, ontologies, and \gls{kg}). The most common methods within this category, which is the focus of this study, are described in detail below. 

	\subsection{Definition Overlap Systems}
		The definition overlap, or Lesk algorithm named after its author, is based on the commonality of words between two sentences, where the first sentence is the context of word $w_t$ and the the second is the definition of a given sense from the knowledge base \cite{lesk1986WSD}. The definition with the highest word overlap is considered the correct sense. However, the Lesk algorithm has major limitations, i.e., being highly sensitive to the exact word match and having a concise definition within WordNet. To overcome this limitation, Nanerjee and Padersen \cite{banerjee2003extended} expanded on Lesk's algorithm to include related concepts within the knowledge base. Related concepts are identified through direct relations with the candidate sense (e.g., hypernyms or meronyms).
		
	\subsection{Semantic Similarity Systems}
		Since the introduction of WordNet, many semantic similarity measures have been developed. Some of the most relevant measures were discussed in \cite{almousa2020exploiting,cai2017hybrid}. This technique follows the intuition that words that appear in a sentence are coherently contextual, and should therefore be highly related within a conceptual knowledge base such as WordNet.  

		Pedersen et al. \cite{pedersen2005maximizing} introduced a variation to the Lesk overlap approach by proposing an exhaustive evaluation of all possible combinations of sentences that can be constructed by all candidate senses within a context window. The context window is the words surrounding a target word. The Pedersen algorithm can be expressed as a general disambiguation framework based on a semantic similarity score. The framework can be described as follows: for a target word $w_i$, $\hat{S}$ is chosen such that it maximizes the sum of the most similar sense with all other words' senses based on the following equation \cite{pedersen2005maximizing,navigli2009word}: 
		\begin{equation}
		\hat{S}=_{S\in Senses(w_i)}^{\arg{max}}\sum_{w_j\in T:w_j\neq w_i}^{s} {_{{S}'\in Senses(w_j)}^{max} score(S,{S}')},
		\end{equation} 
		where $T=(w_1,...,w_n)$ is the set of all words in a text, $Senses(w_i)$ is the full set of senses of $w_i\in T$. The formula measures the contributions of all context words with the most suitable sense. Pedersen's algorithm, as shown in Algorithm \ref{alg:pedersen2005}, can use any semantic similarity measure. However, their results as reported in \cite{pedersen2005maximizing} are much lower than some of the recent approaches of this category, as shown below:
		
		\begin{algorithm}
			\SetKwInOut{Input}{Input}
			\SetKwInOut{Output}{Output}
			\Input{${w_t}$: Target word}
			\Output{${i}$: Index of maximum related sense}
			\ForEach{Sense $s_{ti} \in Senses\_of(w_t)$}{
				Initialize $score_i$ $\leftarrow$ $0$ \\
				\ForEach{word $w_j \in ContextWindow(w_i)=\{w_j:j\neq i\}$} 
				{
					Initialize $maxScore_j$ $\leftarrow$ $0$ \\
					\ForEach{Sense $s_{jk} \in  Senses\_of(w_j)$}{
						\uIf{$maxScore_j<relatedness(s_{ti},s_{jk})$}{
							$maxScore_j=relatedness(s_{ti},s_{jk})$
						}
					}
					\uIf{$maxScore_j>threshold$}{
						$score_i+= maxScore_j$
					}
				}
			}
			Return $i$ such that $score_i \geq score_j, \forall j,1 \leq j \leq n, n= $ number of words in the sentence.
			\caption{Maximum Relatedness Disambiguation \cite{pedersen2005maximizing}} 
			\label{alg:pedersen2005}
		\end{algorithm}
	
		A more recent study conducted by Mittal and Jain \cite{mittal2015word} utilized an average of three semantic similarity measures, some of which include Wu and Palmer ($Sim_{wu}$) measure \cite{wp1994}, Leacock and Chodorow path-based measure ($Sim_{lch}$) \cite{leacock1998combining}, and a node counting distance measure. The average of all three similarity measures is assigned as a similarity value between each sense of an ambiguous word and all neighboring words (context) \cite{mittal2015word}.
	
		\subsection{Heuristic Systems}
		Based on linguistic properties, heuristics are applied to evaluate word senses. The main idea is based on the ranking of sense distribution within a training dataset. Three main heuristic models have been developed to solve the \gls{wsd} task: \gls{mfs}, one sense per discourse, and one sense per collocation.
		\begin{enumerate}
			\item \gls{mfs} is based on the frequency distribution of senses within the training dataset (i.e., SemCor and \gls{omsti}). For a word $w$, the sense with the highest frequency is ranked first $w_s^1$, and the sense with the second highest frequency is ranked second $w_s^2$, and so on. Table \ref{tb:senseFrequencies} depicts the ranking of the noun senses for \textit{`plant'} within SemCor dataset. In fact, senses in WordNet itself are ranked based on their frequency of occurrence in semantic concordance texts\footnote{\url{https://wordnet.princeton.edu/documentation/wndb5wn}} \cite{navigli2009word}.
			\item One sense per discourse argues that the meaning of a word is most likely preserved within a specific text/domain, rather than in general.
			\item One sense per collocation narrows the preservation of meaning within collocation instead of a domain.
		\end{enumerate}
		Once the challenging part of ranking the senses within the knowledge base is complete, disambiguating a word would be as simple as selecting the most frequent sense from the training dataset; which is referred to as \gls{mfs} baseline. The first sense selection from WordNet is also considered a baseline approach. These baseline approaches yield a moderate accuracy between 55.2\% and 67.8\% as reported in SemEval-07 and SemEval-15, respectively \cite{raganato2017word}.
		
		\begin{table}[!t]
			\centering
			\caption{WordNet sense ranking based on SemCor frequencies}
			\label{tb:senseFrequencies}
			\begin{tabular}{llr}
				\hline
				\textbf{Sense} & \textbf{Definition} & \textbf{Frequency} \\
				\hline 
				plant-1 & Buildings for carrying on industrial labor & 338 \\
				plant-2 & A living organism lacking the power of locomotion & 207 \\
				plant-3 & Something planted secretly for discovery by another & 2\\
				plant-4 & An actor situated in the audience whose acting is & 0 \\
				& rehearsed but seems spontaneous to the audience &\\
				\hline
			\end{tabular}
		\end{table}
	\subsection{Graph-based Systems}
		Several other methods exploited the knowledge base structure and attempted to construct a sub-graph to determine the appropriate sense within a sentence. Navigli and Lapata constructed a graph containing all possible combinations of the ambiguous words' senses, where each node of the new graph represents a sense of one of the word sequence, while edges correspond to relationships between senses. Once the graph is constructed, each node is assessed based on the shortest path measure to determine the most suitable sense for each word that provides the highest context \cite{navigli2007graph}.

\subsection{Knowledge-based Benchmarking Systems}\label{subsec:State-of-the-art Knowledge-Based Systems}
	The following are the knowledge-based systems that have been used as a benchmark and will be compared to our system.  
	
	\textbf{Lesk:} The original Lesk algorithm is based on a gloss overlap between the definitions of the ambiguous word's senses and its sentence (i.e., context). The sense definition with the maximum overlap with the word's sentence is selected as the correct sense \cite{lesk1986WSD}. Lesk\textsubscript{ext} is an extension of the original gloss overlap, which extended the gloss to include terms that share one or more relations with the ambiguous term in the \gls{kg}. They also employed the \gls{tfidf} weights to compute the final similarity between the extended gloss and the context \cite{banerjee2003extended}. Finally, Lesk\textsubscript{ext+emb} incorporated \gls{lsa} to select the appropriate sense using semantic vector similarity instead of \gls{tfidf} vector similarity. They re-weighted the terms using an \gls{igf}, viewing all extended glosses as a corpus compared to the \gls{idf} approach. Beyond using the distributional semantic space, the latter overcame the bag of words overlap limitation in the original Lesk algorithm by using a vector cosine similarity \cite{basile2014enhanced}. However, the Lesk algorithm is dependent on the matching of terms between the compared texts. Moreover, the algorithm would fail if the compared text contains synonym terms rather than the exact terms. In addition, none of the overlap approaches take into consideration the sequence of terms within the sentence itself.
	
	\textbf{UKB:} UKB employed a graph-based PageRank approach on the entire WordNet graph, which is a completely different approach from Lesk's. To optimize the PageRank algorithm over WordNet, they constructed a subgraph for a text window (typically a sentence or a few contiguous sentences). The subgraph included the senses of all open-class (ambiguous) terms and the rest of the text as a context \cite{agirre2009personalizing}. 
	An extended version of UKB, namely UKB\textsubscript{gloss}, used extended WordNet to transform the glosses into disambiguated synsets. This implementation of UKB also incorporated sense frequencies to initialize context words \cite{agirre2014random}. 
	The latest release of UKB is UKB\textsubscript{gloss18}, which includes the optimal parameters for the software to guarantee optimal performance. For example, they used a window of over 20 words as a context of each target word and 30 iterations for the personalized PageRank algorithm. They also confirmed that using WordNet versions 1.7.1 and 2.0 resulted in better performance since they match the annotated datasets \cite{agirre2018risk}. Furthermore, the authors highlighted the use of an undirected graph as a limitation for the PageRank algorithm \cite{agirre2014random}.
	
	\textbf{Babelfy:} A graph-based approach integrated entity linking and \gls{wsd} based on random walks with restart algorithm \cite{tong2006fast} over BabelNet, which is an extensive multi-graph semantic network integrating entities from WordNet and Wikipedia or Wiktionary. Babelfy employs the densest subgraph heuristic for selecting the most suitable sense of each text fragment. For a target word, Babelfy considers the entire document instead of the sentence alone \cite{moro2014entity}. This approach is also bound by the PageRank algorithm limitations with respect to WordNet \gls{kg}. 
				
	\textbf{WSD-TM:} This is a graph-based \gls{wsd} system that uses a topic modeling approach based on a variation of the \gls{lda} algorithm. This approach applies the whole document as a context to disambiguate all open-class words within the document. WSD-TM views document as synsets and synset words rather than topics and topic words, then performs the \gls{lda} algorithm based on that assumption \cite{chaplot2018knowledge}. 
				
	\textbf{Baselines:} Senses in WordNet are ranked based on their frequency of occurrence in semantic concordance texts\footnote{\url{https://wordnet.princeton.edu/documentation/wndb5wn}}. Therefore, selecting the first sense of the target word in WordNet is presented as a baseline. Another baseline is based on the \gls{mfs} extracted from the training dataset (SemCor and/or \gls{omsti}). 
	
 \subsection{Critical Analysis of the Related Work}\label{subsec:WSD_Critical Analysis of the Related Work}
	Although the above-mentioned benchmarking systems are all knowledge-based, they can be further classified into three subcategories based on their implemented algorithm. The first subcategory is the definition overlap, the second is the graph-based (i.e., PageRank), while the third is topic modeling. 
	The Lesk systems follow the definition overlap, which limit the similarity between two texts on the term's exact match. Furthermore, the original Lesk algorithm adopts a bag-of-words approach. Although it was enhanced with a vector-based approach in subsequent literature, none of the overlap methods considered the broader context of the document. 
	
	The UKB systems employ a graph-based method (i.e., PageRank). The PageRank algorithm is time-consuming and requires intensive computational power to weigh the links between WordNet concepts. Furthermore, some of these systems employ the Lesk algorithm for the initial weights linking any two concepts \cite{mihalcea2005unsupervised}, while others use a collection of semantic similarity measures including JCN, LCH, and Lesk \cite{sinha2007unsupervised,agirre2014random}. The personalized PageRank optimizes the performance by using a subgraph approach. However, this is achieved at the cost of context reduction, as the optimal results of UKB considers a window size of 20 words, which could span multiple sentences \cite{agirre2018risk}. 
	
	The WSD-TM system relies on the document topic as the main disambiguating context. Despite the importance of the global document context, WSD-TM overlooks the importance of the word's local surroundings, which is considered a local context. Furthermore, this system also employs Lesk similarity to model relationships between synsets as one of its priors to the \gls{lda} algorithm. A major limitation that applies to most systems in these three categories is that they follow a bag-of-words approach, ignoring the sequence of the terms within the sentence, which we believe is a critical factor to disambiguate a word within its sentence and discourse contexts. 
	
\begin{figure}[!ht]
		\centering
		\includegraphics[trim=0 0 0 60,clip,width=.7\columnwidth]{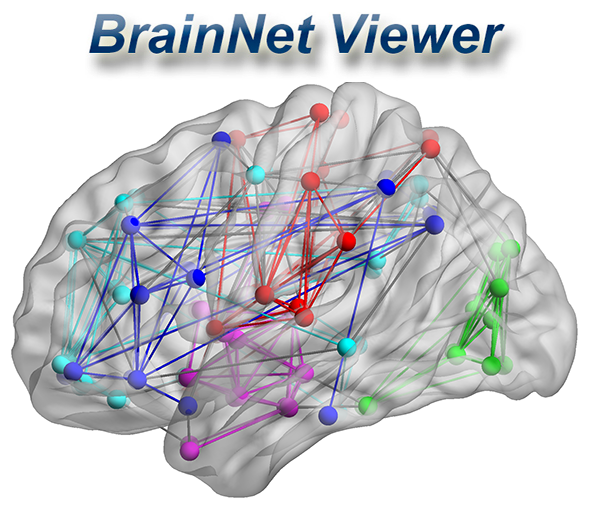}
		\caption[Visualization of the human brain network using the BrainNet viewer]{Visualization of the human brain network using the BrainNet viewer \cite{xia2013brainnet}}
		\label{fig:BrainNetViewer}
	\end{figure}

	Research published in neuroscience journals shows that human brain models suggest that semantic memory is a construction of conceptual knowledge based on a widely-distributed network \cite{patterson2007you}. Based on some models, the brain networks consist of neurons, neuronal populations, or brain regions that can be viewed as nodes, and the structural or functional connectivity viewed as edges linking these nodes together \cite{liao2017small}. Fig. \ref{fig:BrainNetViewer} describes such a network with functional relationships connecting various brain regions (nodes). Furthermore, structural or functional connectivity refers to the anatomical pathways between neurons, neuronal populations, or brain regions, depending on the spatial scales of interest. These structural and functional connections form a biological route for information transfer and communication \cite{patterson2007you,van2013network}. If we compare the \gls{kg} to our brain, viewing concepts as nodes and relations as structural and functional connections, we can rely on widely-distributed \gls{kg} to extract various semantic knowledge, including similarity and relatedness between nodes using the structural and functional relationships, respectively.
	
	Inspired by the brain models, we attempt to overcome the limitations mentioned above as follows: we argue that the sequential connectivity of terms has an essential part in forming the overall context of the sentence. Beneath the sequential connectivity, there exists structural and functional relationships that construct the term's context. These relationships are measured by semantic similarity and relatedness within the \gls{kg}. 
	
	Consider the following two sentences:
	\begin{itemize}
		\item \textit{``John has all his \textbf{faculty} members at the meeting table."}
		\item \textit{``John has all his \textbf{faculties} and could think clearly and logically"}	
	\end{itemize}
	
\begin{figure}[!ht]
		\centering
		\includegraphics[trim=20 490 40 200, clip, width=.7\columnwidth]{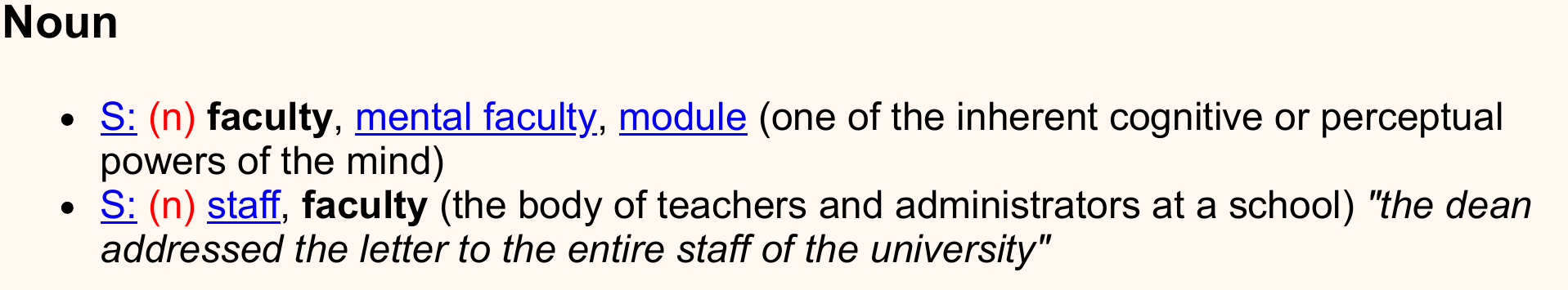}
		\caption{Senses of the word `faculty' in WordNet}
		\label{fig:senses_faculty}
	\end{figure}

	The word \textit{faculty} (lemma of \textit{faculties}) has two distinct meanings (see Fig. \ref{fig:senses_faculty}), and without the rest of the sentence or other external context (e.g., knowing that John is a Dean at a university), it is challenging to distinguish the correct meaning. Since humans use and rely on context to disambiguate words, machines are even more dependent on it.

	If we remove all words that follow \textit{faculty} from both sentences, it will not be easy, as a human being, to understand the correct meaning. This difficulty is derived from the fact that the term \textit{faculty} is ambiguous. However, as we add more context to the sentence, the meaning becomes more evident in each sentence. More importantly, our brain will be able to establish functional connectivities between the terms of the sentence and infer additional knowledge, such as \textit{John} could be working at a university as a Chairperson or a Dean. 
	
\begin{figure}[!t]
		\centering
		\begin{subfigure}[b]{.7\columnwidth}
			\centering
			\includegraphics[trim=20 370 40 200, clip, width=\textwidth]{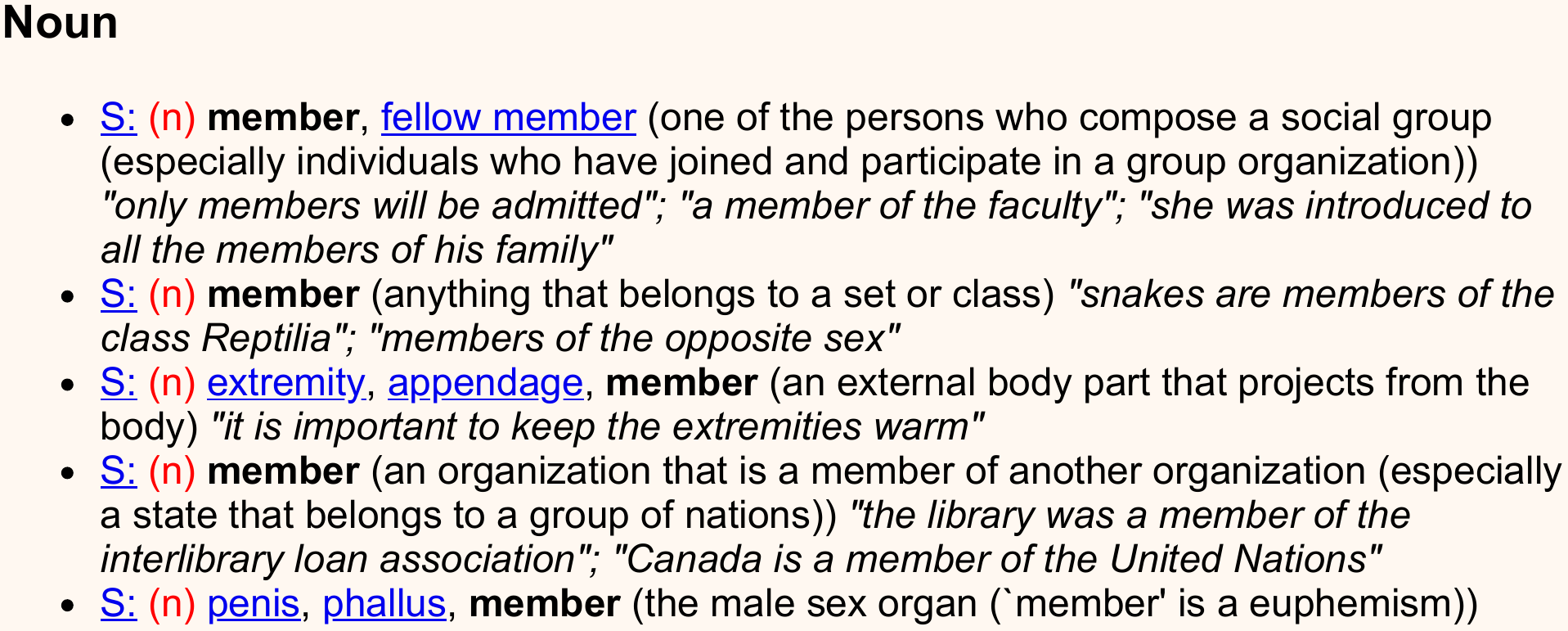}
			\caption{Senses of the word `member' in WordNet}
			\label{figsub:members_senses}
		\end{subfigure}
		\begin{subfigure}[b]{.7\columnwidth}
			\centering
			\includegraphics[trim=20 340 40 200, clip, width=\textwidth]{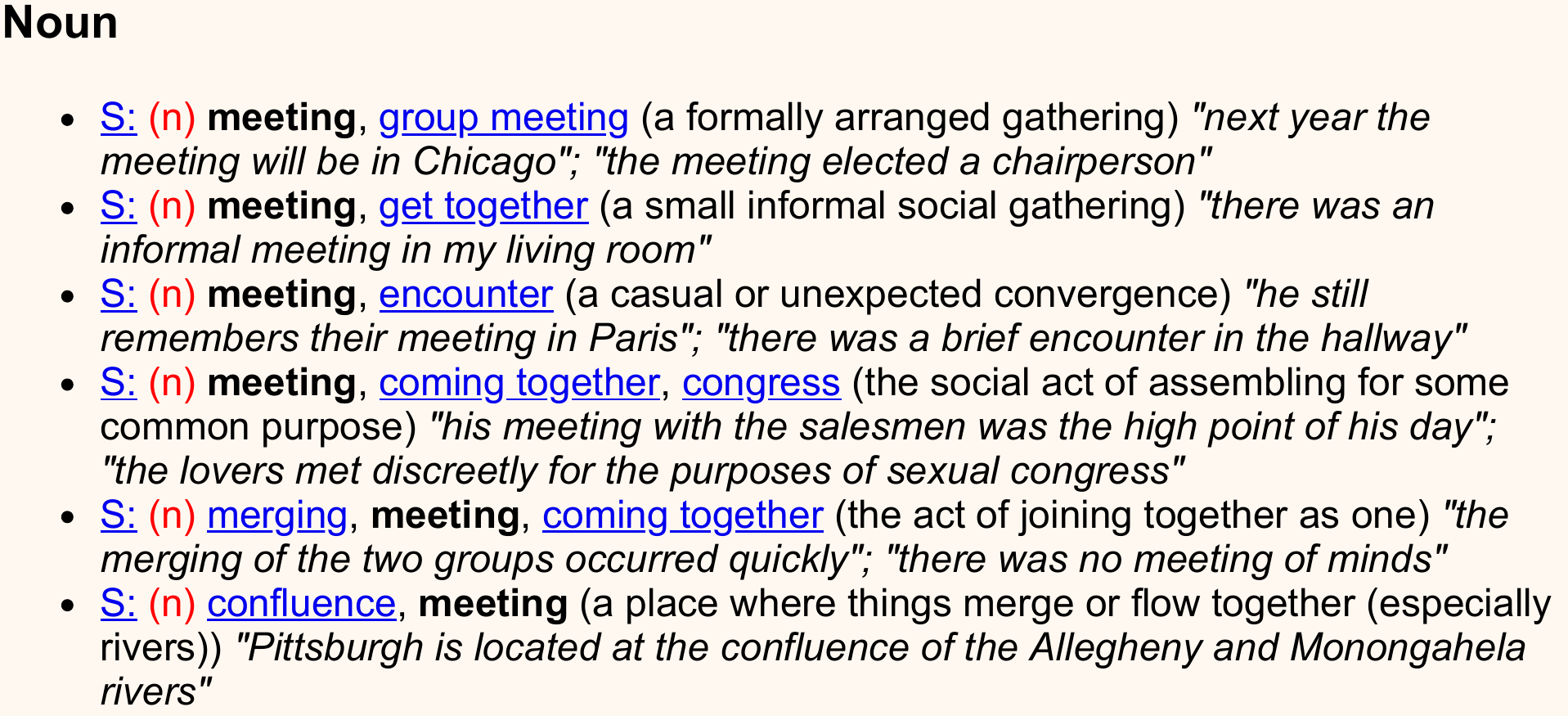}
			\caption{Senses of the word `meeting' in WordNet}
			\label{figsub:meeting_senses}
		\end{subfigure}
		\begin{subfigure}[b]{.7\columnwidth}
			\centering
			\includegraphics[trim=20 380 40 200, clip, width=\textwidth]{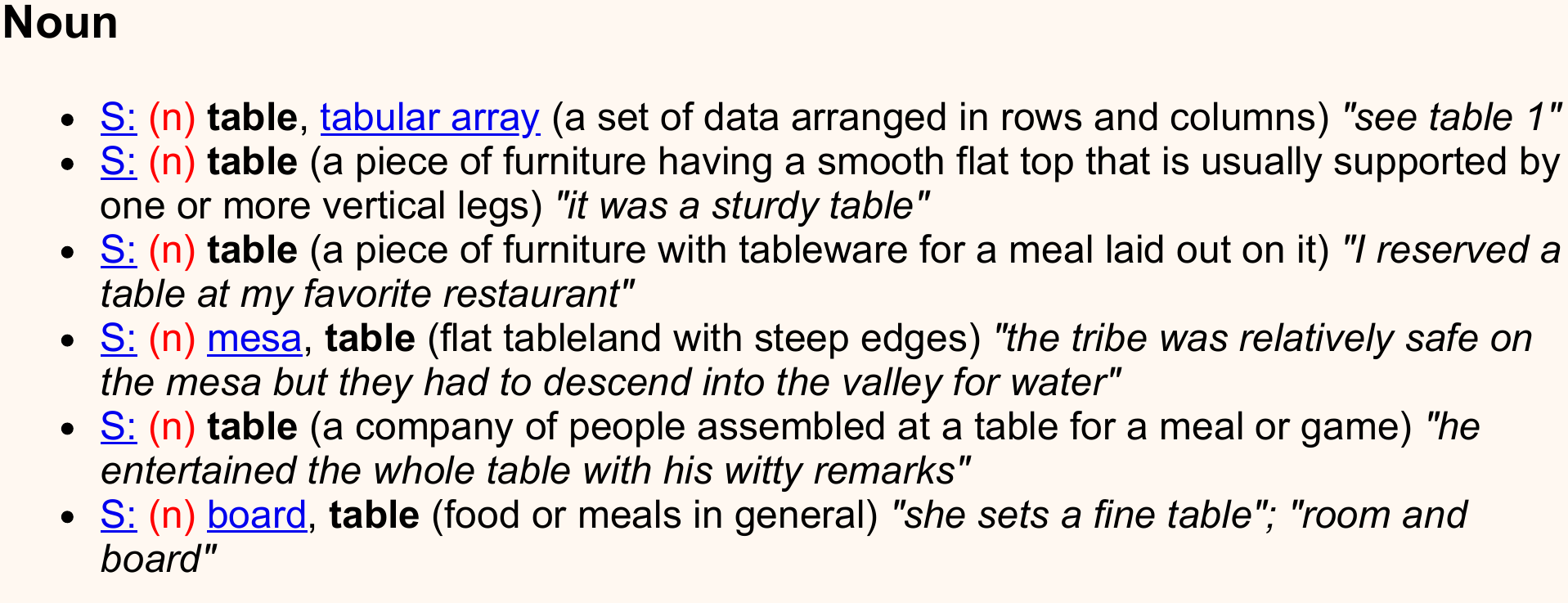}
			\caption{Senses of the word `table' in WordNet}
			\label{figsub:table_senses}
		\end{subfigure}
		\caption[Definitions for the terms `member', `meeting', and `table' in WordNet]{Definitions for the terms `walk' and `bank' in WordNet\footnotemark}
		\label{fig:senses_meeting_members_meeting}
	\end{figure}

	Initially, our brain could not understand the meaning of \textit{faculties} because it could not make the connection between the term and its surrounding context \{`John', `has', `all', `his'\}. However, as soon as the context was enriched with \{`members', `at', `the', `meeting', `table'\}, our brain was able to create a context from the joint meanings of the core terms in the sentence \{`John', `faculty', `member', `meeting', `table'\}, hence, disambiguating the sentence. Surprisingly enough, the three terms \{`member', `meeting', `table'\} are also ambiguous, with even more senses to choose from (refer to Fig. \ref{fig:senses_meeting_members_meeting}). However, our brains can connect the various meanings of each term and determine the context of the full sentence. Our main observation here demonstrates that humans tend to connect terms/things based on the various associations that connect them, in addition to its prior heuristic knowledge about the ambiguous terms. The prior heuristic knowledge is represented by the common use of the terms presented in the sequence.
	
	To summarize, the four points below are essential for disambiguating words within a sentence; hence, we incorporate them into our proposed \gls{wsd} algorithm:
	\begin{itemize}
		\item The sequence of the terms within the sentence;
		\item The connectivity between various concepts (i.e., senses) of ambiguous terms;
		\item Basic heuristic knowledge of each term and its various concepts (i.e., senses);
		\item The broader context of the document.
	\end{itemize}

	\footnotetext{\url{http://wordnetweb.princeton.edu/perl/webwn}}

The limitations in current \gls{wsd} systems motivates us to pursue the following objectives:
\begin{itemize}
	\item Address the limitations in existing knowledge-based \gls{wsd} methods;
	\item Investigate the effect of semantic similarity measures, word sense heuristic, document context, and average sentence size on disambiguating words;
	\item Propose a new algorithm that exploits semantic similarity, word sense heuristic, and document context to solve All-Words \gls{wsd} task;
	\item Evaluate our approaches by using gold-standard benchmarks and state-of-the-art methods to demonstrate their robustness and scalability.
\end{itemize}

To achieve the above-mentioned objectives, we propose a novel \gls{scsmm} algorithm within a comprehensive knowledge-based \gls{wsd} system. Our proposed algorithm follows the disambiguation process of the human brain by exploiting the local context within the sentence, prior knowledge of the term's usage, and the global context of the document, represented by the semantic similarity between terms, terms frequency heuristics, and document context, respectively.

\section{Proposed Method}\label{sec:Proposed Method}
\begin{figure}[!ht]
		\centering
		\includegraphics[width=.90\textwidth]{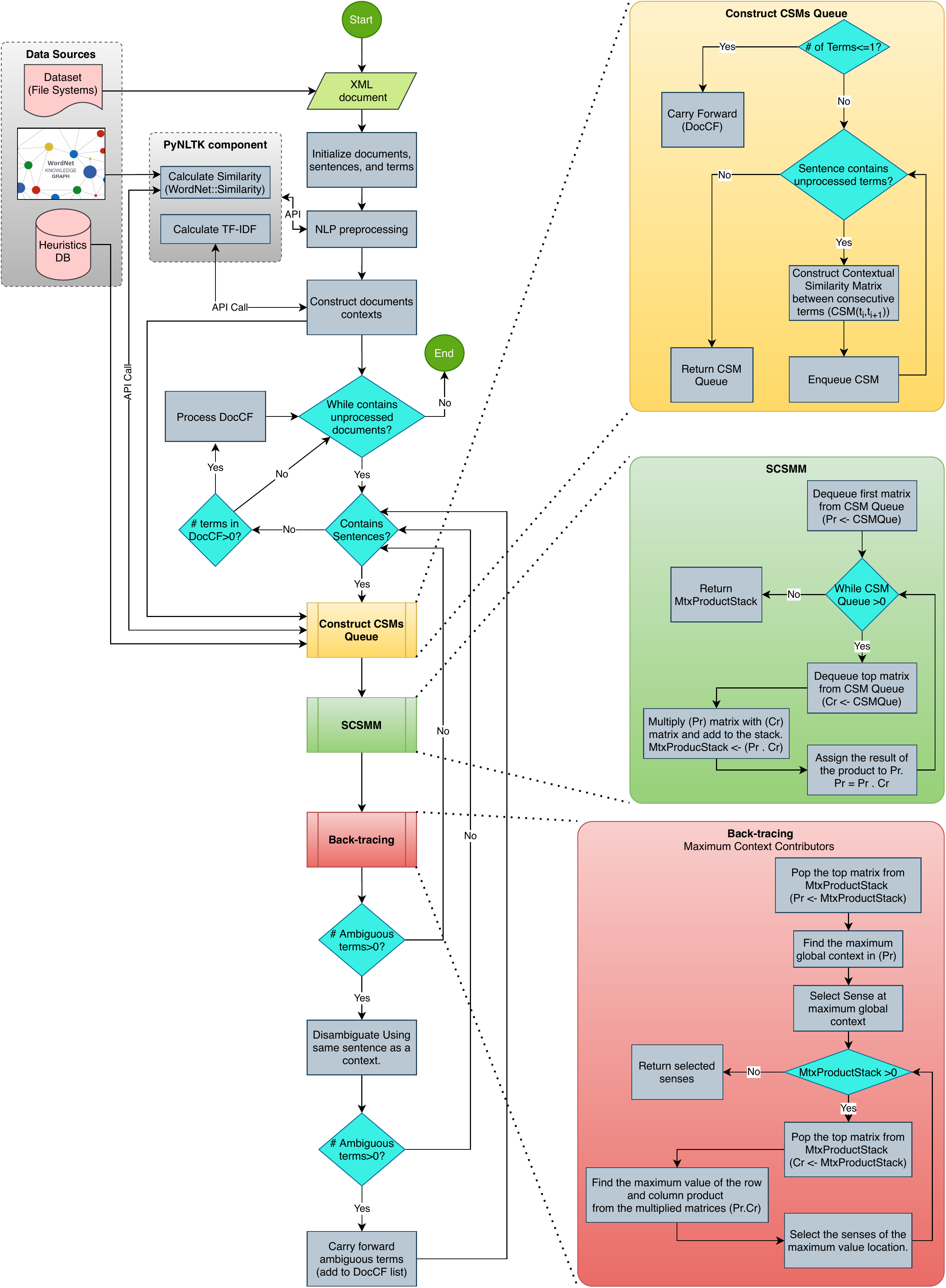}
		\caption{Flowchart for the proposed WSD algorithm}
		\label{fig:WSD_System flowchart}
	\end{figure}

This section presents a novel, context-aware \gls{wsd} algorithm based on a \gls{kg} semantic similarity measure. Our main intuition is derived from the brain's basic steps to analyze and disambiguate words in context (i.e., sentence and document) as described in Section \ref{subsec:WSD_Critical Analysis of the Related Work}. Fig. \ref{fig:WSD_System flowchart} describes the main tasks of the proposed \gls{wsd} method, starting from parsing the XML content of the dataset and the \gls{nlp} preprocessing tasks, followed by the construction of a document's context. The document context consists of all context words within each document (terms with a single sense) that have a nonzero \gls{tfidf} value. The three main \gls{wsd} processes, which make up the \gls{wsd} algorithm, are then executed for each sentence in the document. These include the construction of \gls{csm}s queue, followed by the main \gls{scsmm} algorithm, and finally the identification of the senses that contribute the most to the global context in the back-tracing algorithm. In the cases where any ambiguous terms remain, the carry-forward process is executed to disambiguate them.
	
	The complete \gls{wsd} process, as described in Algorithm \ref{alg:SCSMM}, consists of the \gls{csm} queue construction, a novel \gls{scsmm} and a back-tracing algorithm for an \gls{aw} \gls{wsd} task. The proposed method follows a knowledge-based approach using WordNet as a sense dictionary and the main knowledge resource. Before starting the \gls{wsd} process, standard \gls{nlp} preprocessing steps take place, such as sentence tokenization, stop-words removal, lemmatization, and \gls{pos} tagging. Before delving into the algorithm, the next section presents the core components that construct the \gls{csm}. These are the semantic similarity, sense heuristic, and document context.
							
	\begin{algorithm}[!ht]		
		\SetKwInOut{Input}{Input}
		\SetKwInOut{Output}{Output}
		\Input{${S}$: Sentence with a list of ambiguous words}
		\Output{${\hat{S}}$: Sentence with annotated sense}
		\textbf{Data Structures}:\\
		$CSMQue$: Contextual Similarity Matrices Queue \\
		$MtxProductStack$: A Stack for the produced matrices resulting from the product of consecutive matrices\\
		\For{$i \gets 0$ to $(|TermsOf(S)|-1)$}{
			$CSMQue \xleftarrow{Enqueue}$ Call $ getSemSimMatrix(S_i,S_{i+1})$ \\
		}
		$MtxProductStack \gets $ Call $SCSMM(CSMQue)$\\
		$\hat{S} \gets $ Call $BMCC(MtxProductStack)$
		\caption{WSD Algorithm Using SCSMM}
		\label{alg:SCSMM}	
	\end{algorithm}

	\subsection{CSM Core Components}
	The similarity matrix algorithm described in Algorithm \ref{alg:getSemSimMatrix} employs the aforementioned semantic similarity measure as the similarity measure between the senses of every term and its consecutive term $SCM(t_i,t_{i+1})$. The local context generated by the consecutive terms' similarities is then complemented by the heuristic of each sense and the global context from the document context similarity. As a result, each cell in the \gls{csm} matrix resembles the local context, prior knowledge, and document context (refer to lines 7-9 in Algorithm \ref{alg:getSemSimMatrix}). 
	
	\textbf{1- Semantic Similarity:} 
	A semantic similarity measure represents a direct and local context between consecutive terms. The main idea is to find the maximum pairwise context between senses of the two consecutive terms. However, it is possible to have more than one local context from two words based on the combination of their senses. Various knowledge-based semantic similarity measures have been evaluated in order to determine the best similarity measure for our algorithm. These measures are presented in \cite{almousa2020exploiting}. We further evaluate these measures in Section \ref{subsec:Evaluated Semantic Similarity measures}.
	
		\begin{algorithm}[!b]		
		\SetKwInOut{Input}{Input}
		\SetKwInOut{Output}{Output}
		\Input{
			$Pr_{term}$: First term\\
			$Cr_{term}$: Second term
		}
		\Output{$SimMtx$: Similarity Matrix}
		\textbf{Data Structures}:\\	
		$CSM$: Contextual Similarity Matrix \\
		\textbf{Initialization}:\\	
		$ CSM \gets New Matrix[|Sense(Pr_{term})|][|Sense(Cr_{term})|]\{0\}$ \\			
		\ForEach{$s_i \in Sense(Pr_{term})$}{
			\ForEach{$s_j \in Sense(Cr_{term})$}{
				\tcc{Get the semantic similarity}
				$CSM[i][j] \gets SSR(s_i,s_j)$\\
				\tcc{Apply heuristics as a weighted frequency of each sense}
				$CSM[i][j] *= H(s_i) * H(s_j)$\\
				\tcc{Apply document context similarity of each sense}
				$CSM[i][j] *= DocCtxSim(s_i) * DocCtxSim(s_j)$
			}
		}
		\KwRet{$CSM$}
		\caption{Get Semantic Similarity matrix method}
		\label{alg:getSemSimMatrix}	
	\end{algorithm}
	
	\textbf{2- Sense Heuristic:} In addition to the semantic similarity between senses, each sense has heuristic information that reflects its use frequency. These heuristics are observed from the available training datasets: SemCor and \gls{omsti}. The heuristic function is based on the senses frequency distribution within the training dataset. More formally, for a term $w_i$ that has a set of senses $\{S\}$, and a sense $s_{ij}, 1 \le j \le |S|$, the heuristic function is described as below: 
	
	\begin{equation}
	H(s_{ij})=\left\{
		\begin{matrix}
			P(s_{ij}|w_i) 			& , s_{ij}\in \{S\} \\[6pt]
			\frac{1}{Count(w_i)} 	& , s_{ij}\notin \{S\} \\[6pt]
			1 						& , w_i\notin \{W\}
		\end{matrix}\right.,
	\end{equation}
	where $P(s_{ij}|w_i)$ is the conditional probability of the sense $s_{ij}$ given its term $w_i$, that is computed based on their respective counts within the dataset as follows:
	\begin{equation}
	P(s_{ij}|w_i)=\frac{Count(s_{ij})}{Count(w_i)}
	\end{equation}
	
	Note that if the training dataset does not contain the term $w_i$, its heuristic is set to one, and it will not affect the similarity matrix.
	
	\textbf{3- Document Context:} As described in the semantic similarity, multiple sense-pairs might have high similarity, which indicates various contexts. To determine the appropriate context in the sentence, we crosscheck each sense with the document context obtained from all non-ambiguous terms in the document. Formally, for a given document with sets of ambiguous and non-ambiguous (context) terms $D=\{\{A\} \cup \{C\}\}$, and each ambiguous term $w_i$ $( w_i \in \{A\})$ has a set of senses $\{S_{w_i}\}$, then the sense $s_{ij}$ $( s_{ij} \in {S_{w_i}})$ has a context similarity weight $weight_{CtxD}(s_{ij}|C)$ with the document context $C$ expressed as the average similarity with all context terms $c_k \in \{C\}$ as depicted in the equation below:
	
	\begin{equation}\label{eq:document_Context}
		weight_{CtxD}(s_{ij}|C) = \frac{1}{|C|}\times \sum_{c_k\in C} sim_{jc}(s_{ij},c_k)
	\end{equation}

	\textbf{Illustrative Example:}
	
\begin{figure}[!ht]
		\centering
		\begin{subfigure}[b]{.7\columnwidth}
			\centering
			\includegraphics[trim=20 100 40 400, clip, width=\textwidth]{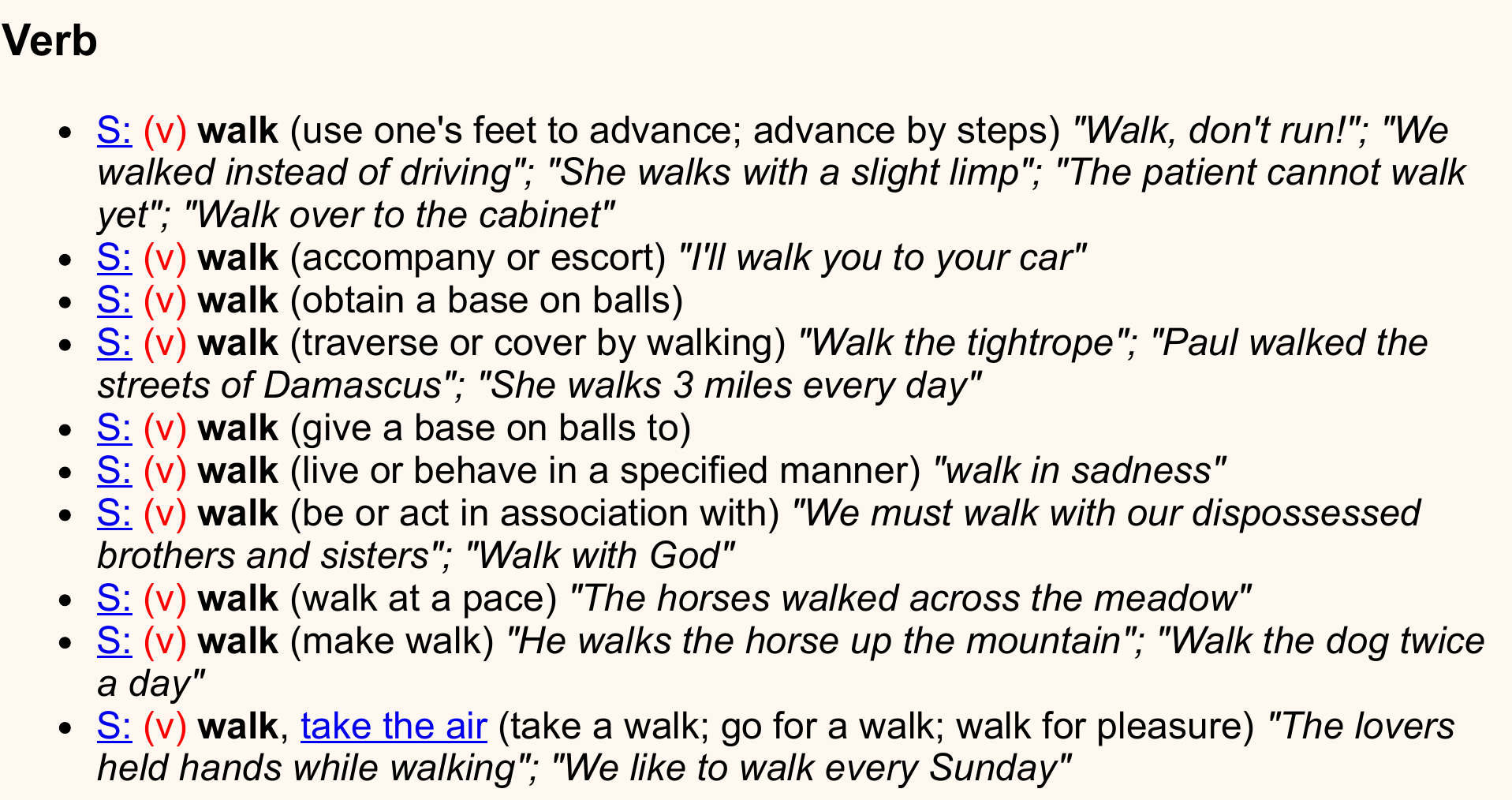}
			\caption{Verb senses for the term walk}
			\label{figsub:wlak_senses}
		\end{subfigure}
		\begin{subfigure}[b]{.7\columnwidth}
			\centering
			\includegraphics[trim=20 420 40 200, clip, width=\textwidth]{Senses_Bank}
			\caption{Noun senses for the term bank}
			\label{figsub:bank_senses}
		\end{subfigure}		
		\caption{Definitions for the terms `walk' and `bank' in WordNet}
		\label{fig:wlak_bank_senses}
	\end{figure}

	Consider the sentence \textit{``I'm \textbf{walking} to the \textbf{bank}"}, with the two ambiguous words \textit{`walk'} and \textit{`bank'}. The similarity matrix (Table \ref{tb:simMatrix_walk_bank_ssm}) shows high similarities between the sense pairs $walk_{v}^{9} - bank_{n}^{3}$, and $walk_{v}^{7} - bank_{n}^{2}$ of 0.092 and 0.077, respectively. These represent a local context for each pair of senses. For more details of these senses and their definitions, refer to Fig. \ref{fig:wlak_bank_senses}.
	 
	\begin{table}[!ht]
		\renewcommand{\tabcolsep}{0.15cm}
		\centering
		\caption{Similarity matrix between terms walk\textsubscript{v} and bank\textsubscript{n}}
		\label{tb:simMatrix_walk_bank_ssm}
		\begin{tabular}{l|ccccccccccc}
			\hline
			& bank1 & bank2 &bank3 &bank4 &bank5 &bank6 &bank7 &bank8 &bank9 &bank10\\
			\hline
			walk1 &0.051&0.053&0.047&0.045&0&0&0&0&0&0 \\
			walk2 &0.048&0.044&0.044&0.037&0&0&0&0&0&0 \\
			walk3 &0.069&0.072&0.063&0.059&0&0&0&0&0&0 \\
			walk4 &0.042&0.039&0.039&0.033&0&0&0&0&0&0 \\
			walk5 &0.069&0.072&0.063&0.059&0&0&0&0&0&0 \\
			walk6 &0.065&0.068&0.060&0.056&0&0&0&0&0&0 \\
			walk7 &0.067&\textbf{0.077}&0.061&0.058&0&0&0&0&0&0 \\
			walk8 &0.066&0.075&0.061&0.058&0&0&0&0&0&0 \\
			walk9 &0.088&0.069&\textbf{0.092}&0.055&0&0&0&0&0&0 \\
			walk10 &0.065&0.063&0.059&0.053&0&0&0&0&0&0 \\
			\hline
		\end{tabular}		
	\end{table}
			 
    For our system to disambiguate such a short sentence with no additional context, it relies only on the semantic similarity. Therefore, the senses $walk_{v}^{9}$ and $bank_{n}^{3}$ would be selected since they have the highest similarity of 0.092 compared to all other combinations. However, when adding heuristics, the results change completely towards another pair $walk_{v}^{1}$ and $bank_{n}^{2}$ with the highest similarity of 0.0236. Intuitively, people would think that the first meaning of walk ($walk_{v}^{1}$) and one of the first two senses of $bank$ would be more meaningful contexts than the rest. This intuition is clearly visible in Table \ref{tb:simMatrix_walk_bank_heuristics} with the top two senses of $bank$ ($bank_{n}^{1}$ and $bank_{n}^{2}$). Note that the heuristic weights for $walk_{v}^{1}$ is $0.9$, and for $bank_{n}^{1}$ and $bank_{n}^{2}$ are $0.35$ and $0.5$, respectively. Heuristics were computed using both of SemCor and \gls{omsti} datasets.
	 
	\begin{table}[!ht]
	 	\centering
	 	\caption{Similarity matrix with heuristics between terms walk\textsubscript{v} and bank\textsubscript{n}}
	 	\label{tb:simMatrix_walk_bank_heuristics}
	 	\begin{tabular}{l|cccc}
	 		\hline
	 		& bank1 & bank2 &bank3 &bank4 \\
	 		\hline
	 		walk1 &0.0158&\textbf{0.0236}&0.0021&0.0010 \\
	 		walk2 &0.0003&0.0004&0.0000&0.0000 \\
	 		walk3 &0.0004&0.0006&0.0001&0.0000 \\
	 		walk4 &0.0001&0.0001&0.0000&0.0000 \\
	 		walk5 &0.0001&0.0002&0.0000&0.0000 \\
	 		walk6 &0.0001&0.0002&0.0000&0.0000 \\
	 		walk7 &0.0001&0.0002&0.0000&0.0000 \\
	 		walk8 &0.0001&0.0002&0.0000&0.0000 \\
	 		walk9 &0.0002&0.0002&0.0000&0.0000 \\
	 		walk10 &0.0001&0.0002&0.0000&0.0000 \\				
	 		\hline
	 	\end{tabular}		
	\end{table}

	Finally, if we are provided with additional context about the sentence, such as non-ambiguous terms within the same document (i.e., river), our brain will shift towards a more concrete context based on the document's main topic, and so does our system. The first sense will have higher similarity than the second one, with the first sense $walk_{v}^{1}$ of 0.153 and 0.151, respectively. The final correct senses in this case would be $walk_{v}^{1}$ and $bank_{n}^{1}$. On the other hand, if the document contained more financial terms (i.e., central\_bank), the other sense would be selected. Based on the above, we employed the document's context similarity, which improves the overall similarity between the senses.

	\subsection{Sequential Contextual Similarity Matrix Multiplication Algorithm}
\begin{figure}[!ht]
		\centering
		\includegraphics[trim=140 55 140 55,clip,width=.7\columnwidth]{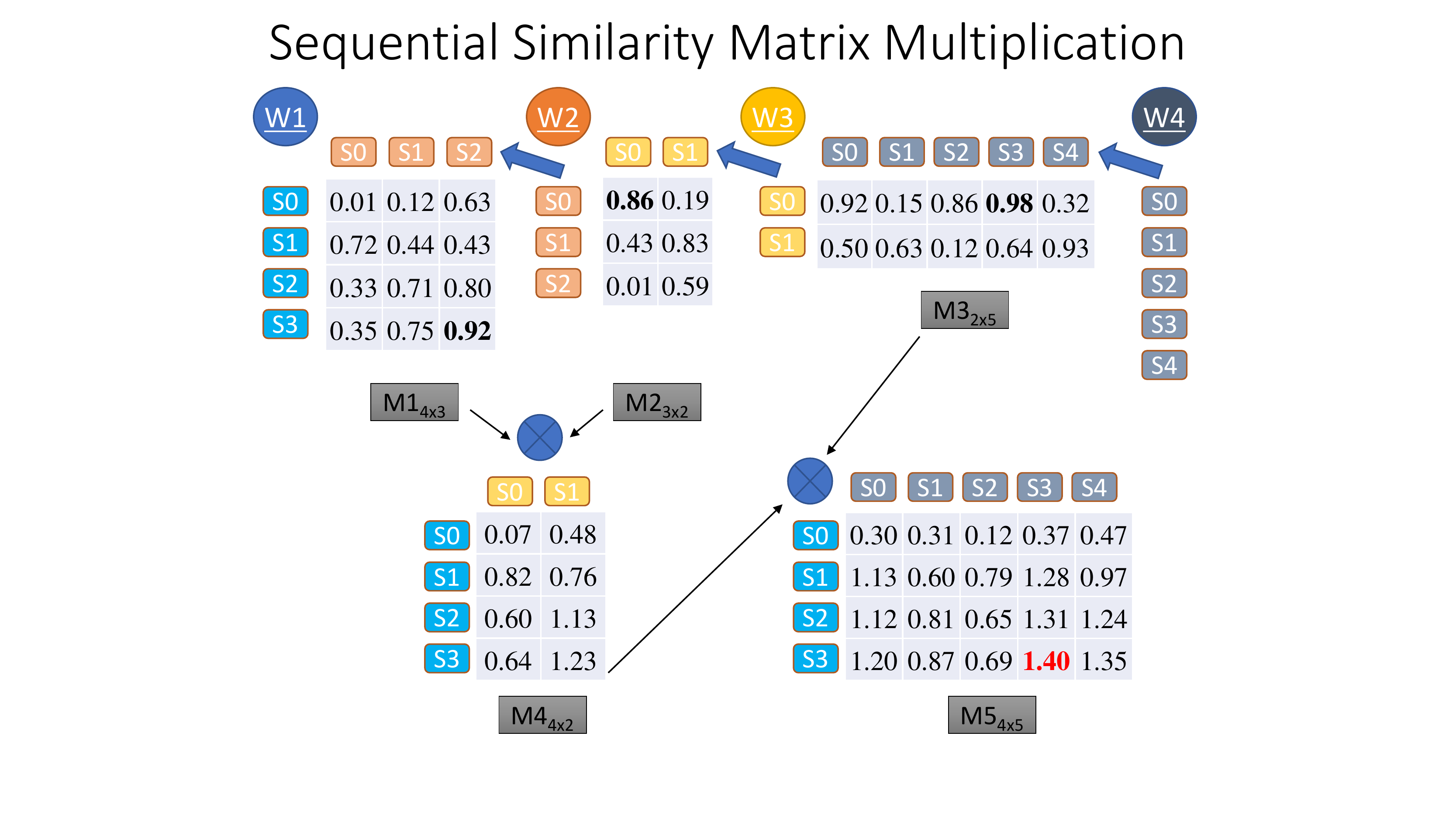}
		\caption{SCSMM illustration}
		\label{fig:SCSMM}
	\end{figure}

	Once all \gls{csm}s are constructed for the sentence, the \gls{wsd} algorithm starts by building a similarity matrix queue ($CSMQue$) from all \gls{csm}s, maintaining their sequence (refer to Algorithm \ref{alg:SCSMM} lines 4-5). Line 6 in the algorithm generates the final matrix based on the sequential multiplication of the matrices, as presented in the \gls{scsmm} algorithm (Algorithm \ref{alg:MatrixMultip}). Fig. \ref{fig:SCSMM} illustrates the sequential multiplication process of the consecutive local \gls{csm}s for a sample sentence with four ambiguous words. Finally, the algorithm applies a back-tracing process to determine the most contributing senses to the maximum global context. Next, we describe the \gls{scsmm} algorithm in detail, followed by the back-tracing algorithm.
	
	\begin{algorithm}[!ht]		
		\SetKwInOut{Input}{Input}
		\SetKwInOut{Output}{Output}
		\Input{$CSMQue$: Contextual Similarity Matrices Queue}
		\Output{$MtxProductStack$: A Stack stores the product of the consecutive matricides}
		\textbf{Data Structures}:\\	
		$Pr_{matrix}$: Stores the previous matrix\\
		$Cr_{matrix}$: Stores the current matrix\\
		$MtxProductStack$: A Stack stores the product of the consecutive matricides\\  
		\textbf{Initialization}:\\	
		$Pr_{matrix} \xleftarrow{Dequeue} CSMQue$\\
		$MtxProductStack \xleftarrow{Push} Pr_{matrix}$ \\ 
		\While{$CSMQue \neq Empty$}{
			$Cr_{matrix} \xleftarrow{Dequeue} CSMQue$ \\
			$MRes \gets Pr_{matrix} \cdot Cr_{matrix}$ \\
			$MtxProductStack \xleftarrow{Push} MRes$ \\ 
			$Pr_{matrix} \gets Cr_{matrix}$
		}
		\KwResult{$MtxProductStack$}
		\caption{Sequential Contextual Similarity Matrix Multiplication}
		\label{alg:MatrixMultip}	
	\end{algorithm}

	\textbf{Similarity Matrices Multiplication:}
	Once all \gls{csm}s are constructed between consecutive terms (see Fig. \ref{fig:SCSMM}, matrices M1, M2, and M3), the matrix multiplication algorithm (Algorithm \ref{alg:MatrixMultip}) starts by multiplying M1 and M2, and the resulting matrix M4 is then multiplied by M3, and so on. The sequential multiplication of matrices guarantees a global context across all words within the sentence. It also guarantees the maximum context value while maintaining the order of the terms within the sentence. The order of words in a sentence is critical to better understand and disambiguate the sentence. Finally, starting with the latest produced matrix, the back-tracing algorithm traces back all senses that contributed to the maximum global context. 
	
\begin{figure}[!ht]
		\centering
		\includegraphics[trim=140 55 140 55,clip,width=.7\columnwidth]{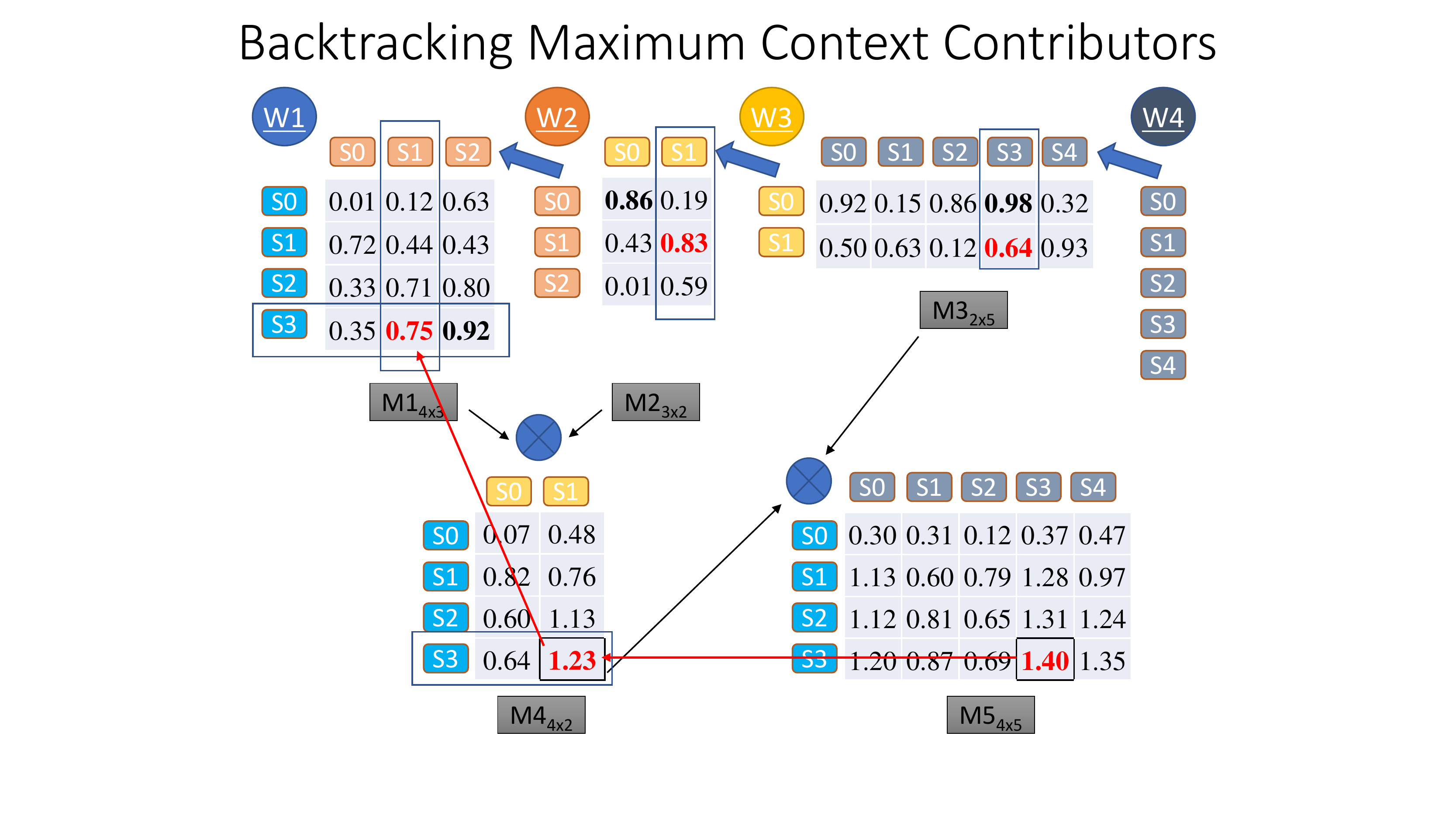}
		\caption{SCSMM back-tracing illustration}
		\label{fig:SCSMM_backtracing}
	\end{figure}

\begin{figure}[!ht]
		\centering
		\includegraphics[trim=120 55 70 55,clip,width=.7\columnwidth]{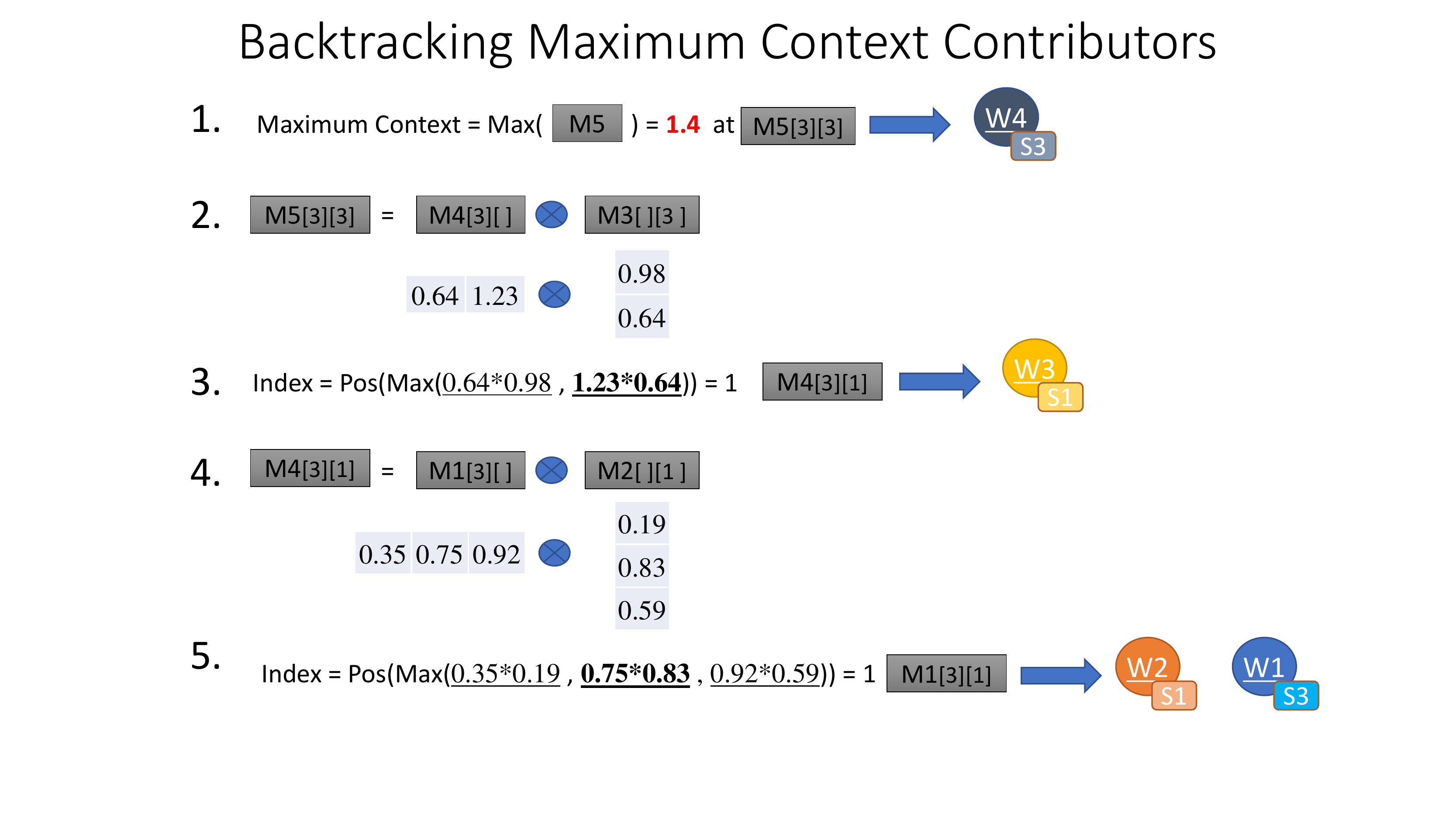}
		\caption{SCSMM back-tracing steps}
		\label{fig:SCSMM_backtracingSteps}
	\end{figure}

	\textbf{Back-tracing Senses:}
	The final step of the \gls{scsmm} algorithm is the back-tracing stage (Algorithm \ref{alg:SCSMM}, line 7). In this stage, we identify the most contributing sense to the sentence's global context (Algorithm \ref{alg:backtracingSteps}). Fig. \ref{fig:SCSMM_backtracing} and \ref{fig:SCSMM_backtracingSteps} illustrate the back-tracing stage as follows: back-tracing begins by selecting the maximum value from the final produced matrix. This value represents the maximum contextual weight for a given sentence. This value is then decomposed into its row and column vectors from the previous matrix multiplication. In step three, we select senses with the maximum product. These are senses that contributed the most to the global context. Finally, steps two and three repeat until no elements are left to decompose. 
	
	As described above, our algorithm is intuitive and its results are explicable. It begins with a local context and then improves the context with heuristics and document context. Finally, it selects the most appropriate sense that contributes to the maximum global context while maintaining terms order.
	
	\begin{algorithm}[!ht]		
		\SetKwInOut{Input}{Input}
		\SetKwInOut{Output}{Output}
		\Input{$MtxProductStack$: A Stack stores the product of the consecutive matricides}
		\Output{$SensesList$: A stock of list of selected Senses}
		\textbf{Data Structures}:\\	
		$Pr_{matrix}$: Stores the previous matrix\\
		$Cr_{matrix}$: Stores the current matrix\\
		$location<r,c,val>$: triple $<$row, col, value$>$ of the location of maximum value in the matrix\\
		\textbf{Initialization}:\\	
		$Pr_{matrix} \xleftarrow{Pop} MtxProductStack$\\
		$location\{r,c,val\} \gets Max(Pr_{matrix})$ \\
		\While{$MtxProductStack \neq Empty$}{	
			$SensesList \xleftarrow{Push} Sense(c)$ \\	
			$Cr_{matrix} \xleftarrow{Pop} MtxProductStack$\\
			\tcc{The index of column that contributes the most to the context}
			$c \gets Max(\{Row_{Cr} . Col_{Pr}\})$ \\
			$location \gets \{r,c,val\} $\\
			$Pr_{matrix} \gets Cr_{matrix}$
		}
		$SensesList \xleftarrow{Push} Sense(c)$ \\	
		$SensesList \xleftarrow{Push} Sense(r)$ \\
		\KwRet{$SensesList$}
		\caption{Back-tracing the maximum context contributing senses}
		\label{alg:backtracingSteps}	
	\end{algorithm}
	
	\subsection{Document Carry-forward Terms:}
	In a few cases, our algorithm is unable to disambiguate a term using the \gls{scsmm} algorithm. This would happen where a term has no local context (zero similarity) with its surrounding terms. In such cases, we first attempt to disambiguate the term using its sentence as a context, including all recently-disambiguated terms. We then select the sense with the maximum similarity with the sentence context. However, if a term could still not be disambiguated within its own sentence, the term is then carried forward to be disambiguated after the entire document is processed. These terms are referred to as \gls{doccf} terms, which are processed after all sentences have been disambiguated to provide a maximum context for these terms. For each \gls{doccf} term, the sense with the maximum average similarity with all terms in the document is selected. 
	
\section{Evaluation and Experimental Results}\label{sec:Evaluation and Experimental Results}
\subsection{Experimental Setup}
	We compared the results of our proposed \gls{scsmm}-\gls{wsd} approach to the state-of-the-art systems based on well-known evaluation datasets. We also employed the commonly used training dataset in this field to obtain sense heuristic. We also compared our approach to the baseline approaches represented by selecting the first sense in WordNet and the \gls{mfs} using both training datasets. To obtain heuristics, we retrieved the senses' annotations from the SemCor and \gls{omsti} training datasets (see Section \ref{subsubsec:Training Datasets}). The SemCor annotations are available as part of the SemCor installation package in the `cntlist' file, and the \gls{omsti} annotations were preloaded to the SQL database from the `keys' file downloaded from \cite{raganato2017word}\footnote{\url{http://lcl.uniroma1.it/wsdeval/home}}.
	
	\subsubsection{Training Datasets}\label{subsubsec:Training Datasets}
	The two large sense-annotated corpora (SemCor and \gls{omsti}) have been used in many supervised approaches for training their models. Both datasets are tagged with WordNet senses; one of which is manually annotated, while the other is automatic. 
	\begin{itemize}
		\item \textbf{SemCor} \cite{miller1994SemCor}: a manually-annotated corpus extracted from the original Brown corpus. The dataset is annotated with \gls{pos}, lemmas, and word senses based on WordNet \gls{kg}. SemCor consists of 352 documents: 186 documents include tags for all \gls{pos} words (nouns, verbs, adjectives, and adverbs), while the remaining 166 contain tags only for verbs. The total number of sense annotations in all documents is 226,040. To our knowledge, SemCor is the largest manually-annotated corpus with WordNet senses, and is the main corpus used in various literature to train supervised \gls{wsd} systems \cite{agirre2010semeval,zhong2010makes}.
		\item \textbf{OMSTI} \cite{taghipour2015OMSTI}: an automatically-annotated corpus with senses from WordNet 3.0. As the name suggests, it contains one million sense-annotated instances. To automatically tag senses, \gls{omsti} used an English-Chinese parallel corpus\footnote{\url{http://www.euromatrixplus.net/multi-un/}} with an alignment-based \gls{wsd} approach \cite{chan2005scaling}. \gls{omsti} has already shown its potential as a training corpus by improving the performance of supervised systems \cite{taghipour2015OMSTI,iacobacci2016embeddings}.
	\end{itemize}
	
	\subsubsection{Evaluation Datasets (Gold Standard)}\label{subsubsec:Evaluation Datasets (Gold Standard}
	A comprehensive evaluation framework has been presented in \cite{raganato2017word} with the integration of the primary WSD datasets. These datasets were presented as part of the SemEval International Workshop on Semantic Evaluation\footnote{Current workshop website: \url{http://alt.qcri.org/semeval2020/}} between 2002 and 2015. The framework included datasets from five main competitions, as presented in Table \ref{tb:SensEval-SemEval_Goldstandard}.
	\begin{table}[!hb]
		\centering
		\caption{SensEval/SemEval evaluation datasets}
		\label{tb:SensEval-SemEval_Goldstandard}
		\begin{tabular}{llccccc}
			\hline
			Dataset 		& Task 		&  \multicolumn{5}{c}{\# of Senses} 	\\
			\cline{3-7}
			Name			& Method	& NN 	& V 	& Adj 	& Adv 	& Total \\
			\hline
			SensEval2 (SE2) \cite{edmonds2001senseval2} 	& LS, AW	& 1066	& 517	& 445	& 254	& 2282	\\
			SensEval3 (SE3) \cite{snyder2004senseval3}	& LS, AW	& 900	& 588	& 350	& 12	& 1850 	\\ 
			SemEval-07 (SE07) \cite{pradhan2007semeval07}	& LS		& 159	& 296	&	-	&	-	& 455 	\\
			SemEval-13 (SE13) \cite{navigli2013semeval13}	& LS, AW	& 1644	&  -	&	-	&	-	& 1644 	\\
			SemEval-15 (SE15) \cite{moro2015semeval15}	& LS, AW	& 531	& 251	& 160	& 80	& 1022 	\\
		\end{tabular}		
	\end{table}
	
	We further analyzed the datasets to determine the average sentence size, context size, and ambiguity rate within each dataset. Table \ref{tb:wsd_GD_Stats} depicts the statistics for each dataset. The average sentence size is calculated based on the number of annotated terms/processed sentence. Sentences that do not contain any terms are omitted. The context size is measured by the number of terms that have a single sense, hence unambiguous terms. Finally, the percentage of ambiguity is computed based on the number of ambiguous terms to the total number of terms. For example, \textit{SemEval-07} has the highest ambiguity rate of 94\%, with only 26 out of 455 terms that are not ambiguous (only one sense), and the smallest average sentence size with an average of only three terms/sentence. Note that the ambiguity rate is inversely correlated with context size, which could degrade the disambiguation score, as presented in the results in Section \ref{subsec:ch4_Experimental Results Performance Analysis}.

	\begin{table}[!ht]
		\centering
		\caption{Statistics of WSD gold standard dataset}
		\label{tb:wsd_GD_Stats}
		\begin{tabular}{lccccc}
			\hline
			Criteria		&SE2	&SE3	&SE07	&SE13	&SE15\\
			\hline
			\#Doc			&	3	&	3	&	3	&	13	&	4	\\
			\#Sent*			&	242	&	297	&	120	&	301	&	133	\\
			\#Terms			&2282	&1850	&	455	&	1644&	1022\\
			AvgSentSize		&	9	&	6	&	3	&	5	&	7	\\
			Single sense	&	442	&311	&	26	&	348	&189\\
			Ambiguity rate	&81\%	&83\%	&94\%	&79\%	&82\% \\
			\hline
		\end{tabular}	
	\end{table}
	
	Furthermore, out of those ambiguous terms, Table \ref{tb:Goldstandard_Ambig_stats} depicts the granularity level for each \gls{pos} on all datasets combined. The granularity level reflects the average number of senses for each term, and negatively impacts disambiguation performance. Having a high granularity level makes the disambiguation decision very difficult even for humans, explaining the relatively low inter-agreement score between annotators. The annotators' inter-agreement score ranges between 72\% to 80\% on \gls{aw} task. The average granularity level for verbs is the highest compared to all other \gls{pos}; on average, each verb term has 10.95 senses compared to 5.71, 4.7, and 4.4 senses for the nouns, adjectives, and adverbs, respectively. The fourth row presents the maximum number of senses within each \gls{pos}, where the maximum number of senses in verbs reaches up to 59, compared to 33, 21, and 13 senses for the nouns, adjectives, and adverbs, respectively. Both nouns and verbs are highly granular, explaining most systems' results as will be described in Section \ref{subsec:ch4_Experimental Results Performance Analysis}. 
	The mode and median also explain the results in Section \ref{subsec:ch4_Experimental Results Performance Analysis}, as most ambiguous verbs have four senses compared to two senses in all other \gls{pos}.
	
	\begin{table}[!ht]
		\centering
		\caption{Ambiguous terms statistics for all gold standard datasets}
		\label{tb:Goldstandard_Ambig_stats}		
		\begin{tabular}{lcccc}
			\hline
			&Noun	&Verb	&Adjective	&Adverb	\\
			\hline
			\# of terms	&4300	&1652&	955	&346\\
			\# of ambiguous	&3442&	1555&	732	&208\\
			Average granularity&5.7	&11.0&	4.7	&4.4\\
			Max \#senses&	33&	59&	21	&13\\
			Mode	&	2&	4&	2&	2\\
			Median	&	5&	7&	4&	3\\				
			\hline
		\end{tabular}		
	\end{table}

\subsection{Evaluation Metric}\label{subsec:ch4_Evaluation Metrics}
	Three main metrics are used to evaluate any \gls{wsd} system performance: Precision, Recall, and F1-score. These measures are commonly used in the \gls{ir} field. Assuming, within a dataset, there is a set of manually annotated test words $T=(w_1,...,w_n)$, and for any system, the set of all evaluated/retrieved words is represented as $E=(w_1,...,w_k) : k<=n$, and the set of correctly evaluated words $C=(w_1,...,w_m):m<=k$. Then we can evaluate the system as follow:
	\begin{itemize}
		\item \textbf{Precision:} the percentage of correctly identified words given by the system: 
		\begin{equation}
		    P=\frac{Number\, of\, correct\, words}{Number\, of\, evaluated\, words}=\frac{m}{k},
		\end{equation}
		where $k=|E|$ the total number of evaluated words, and $m=|C|$ the total number of correctly evaluated words.
		\item \textbf{Recall:} the percentage of correctly identified words given by the system out of all test words in the dataset:
		\begin{equation}
		    R=\frac{Number\, of\, correct\, words}{Number\, of\, test\, words}=\frac{m}{n},
		\end{equation}
		where $n=|T|$ the total number of evaluated words, and $m=|C|$ the total number of correctly evaluated words. If a system is able to evaluate every test word in $T$, then, we can say that the system has a $100\%$ coverage; hence, $P=R$. 
		\item \textbf{F1-score:} is a balanced $F_\alpha\mbox{-}score$ where $\alpha=0.5$. The $F_1\mbox{-}score$ is given by the following equation:
		\begin{equation}
		    F_1\mbox{-}score=\frac{2PR}{P+R}
		\end{equation}
		The general $F_\alpha\mbox{-}score$ measures the trade-off between the precision and recall as follows:
		\begin{equation}
		    F_\alpha\mbox{-}score = \frac{1}{\alpha\frac{1}{P}+(1-\alpha)\frac{1}{R}}
		\end{equation}		
	\end{itemize}
			
\subsection{Evaluated Semantic Similarity measures}\label{subsec:Evaluated Semantic Similarity measures}
	In this section, we present various semantic similarity measures that have been evaluated in our experiment. The similarity measure with the best performance is employed to construct the similarity matrix for our algorithm, as shown in Algorithm \ref{alg:getSemSimMatrix}, Line 7. These measures have been discussed in detail in \cite{almousa2020exploiting}. Table \ref{tb:ssr_evaluation_top4} depicts the performance of the top four measures (\textit{LCH, WUP, JCN, and PATH}) on all dataset. As shown in these results, the JCN measure provides the best \gls{wsd} performance across all datasets. The only exception is on \textit{SemEval2013} where both \textit{PATH} and \textit{LCH} outperformed \textit{JCN}. However, using the combined datasets, \textit{JCN} outperformed all other methods. Hence, it is the measure used in our \gls{scsmm} algorithm.
	
	\begin{table}[!ht]
		\renewcommand{\tabcolsep}{0.25cm}
		\centering
		\caption{F1-score for top four semantic similarity methods}
		\label{tb:ssr_evaluation_top4}
		\begin{tabular}{lcccccc}
			\hline
			SSR		&	 SE2	& SE3 		& SE07 		& SE13 		& SE15 		& All\\
			\hline
			LCH		&	72.51	&	69.89	&	61.01	&	64.42	&	66.29	&	67.67 \\
			WUP		&	73.17	&	68.78	&	62.89	&	63.56	&	66.48	&	67.37 \\
			JCN		&	\textbf{78.14}	&	\textbf{72.67}	&	\textbf{64.78}	&	63.44	&	\textbf{68.38}	&	\textbf{69.67} \\
			PATH	&	73.17	&	70.11	&	61.01	&	\textbf{64.66}	&	66.29	&	67.98  \\
			\hline
		\end{tabular}	
	\end{table}		

\subsection{Implementation}\label{subsec:ch4_Implementation}
\begin{figure}[!ht]
		\centering
		\includegraphics[width=.7\columnwidth]{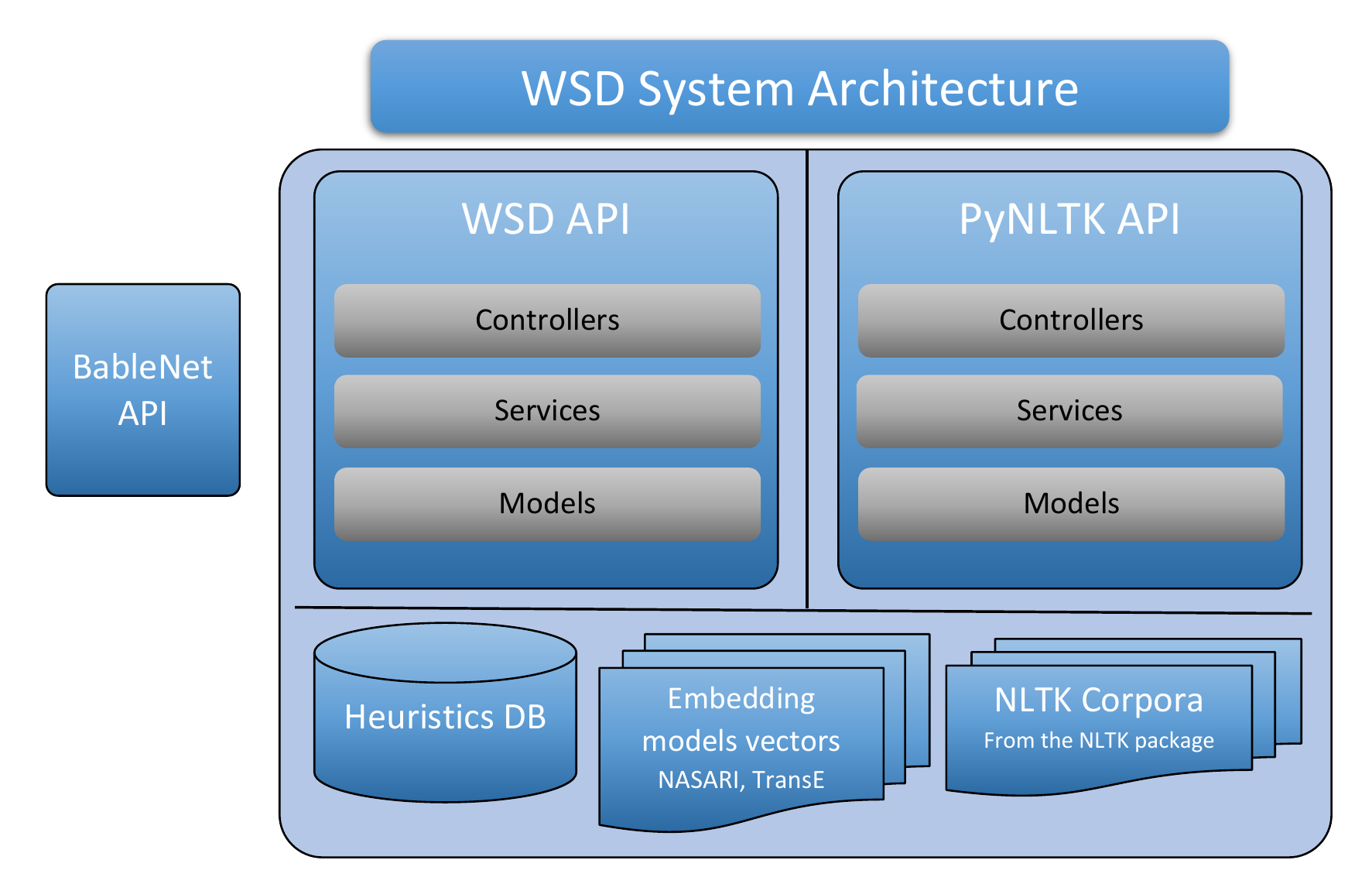}
		\caption{WSD system architecture}
		\label{fig:WSD_System Architecture}
	\end{figure}

	Fig. \ref{fig:WSD_System Architecture} describes the architecture for the proposed \gls{wsd} system. The \gls{wsd} system is built based on the Web API architecture, which includes controllers and models. We further extend the architecture to provide a separate services component that handles the main \gls{wsd} system logic. The architecture consists of two Web API systems: \textbf{WSD API} and \textbf{PyNLTK API}. \textbf{WSD API} is responsible for the core \gls{wsd} algorithm, while \textbf{PyNLTK API} carries out any \gls{nlp} processing tasks, including gloss-based similarity (i.e, Lesk).
	
	The main \gls{wsd} application is a C\# Web API application with three separate layers: controllers, services, and models. The controllers handle the API routing process and trigger the appropriate system logic from the services layer. In return, the services component is responsible for implementing the core \gls{wsd} algorithm. It also connects with the models to add, retrieve, and update data from the database. Furthermore, the services layer is also responsible for establishing any internal or external API calls such as the calls to the \textbf{PyNLTK API} to perform any \gls{nlp} pre-processing required, or the calls the \textbf{BabelNet API}\footnote{\url{https://babelnet.org/}} to obtain BabeleNet synsets, which is required for the NSARI embedding evaluation. 
	
	The second \textbf{PyNLTK API} application is a python-based implementation. The main role of this component is to compute text-based similarity measures such as the \textit{LESK} similarity.
	
	The data is retrieved from three distinct sources. The first is an SQL server database that stores the heuristics datasets (SemCor and \gls{omsti}). The second consists of filesystems that contains pre-calculated embedding vectors for WordNet \gls{kg} from two embedding models: NASARI \cite{camacho2016nasari}, and TransE from \cite{yu2019pykg2vec}. The last is the \gls{nltk} corpora as part of the NLTK\footnote{\url{https://www.nltk.org/}} package. We employed the Brown and SemCor corpora to compute concepts' \gls{ic}.
	
	\begin{table}[!ht]
		\centering
		\caption{Configuration parameters for the SCSMM system}
		\label{tb:configuration_parameters}
			\begin{tabular}{lll}
				\hline
				Name & Description & Best Config. \\
				\hline
				SSR 	& Semantic similarity measure & JCN \\ 
				H(x) 	& Heuristic dataset used s=SemCor, so=SemCor+\gls{omsti} & H(s) \\
				DocCtx	& Document context used in the \gls{csm} flag	& True \\
				DocCF	& Document carry-forward flag	& True \\
				POS\_Of\_Int & The list of \gls{pos} of interest that are being processed & \{n,v,adj,adv\} \\
				DocCtxPOS	& The list of \gls{pos} used as in the document context & \{n,v\} \\
				\hline
			\end{tabular}
	\end{table}
	
	Table \ref{tb:configuration_parameters} outlines the main parameters that control our system, where the right most column shows the optimal configuration that leads to optimal performance. Note that we include all \gls{pos} in the evaluation for the POS\_Of\_Int. However, since adjectives and adverbs merely describe nouns and verbs, respectively, they are not considered a context in DocCtxPOS parameter.
			
\subsection{Experimental Results and Performance Analysis}\label{subsec:ch4_Experimental Results Performance Analysis}
	To validate the robustness of the proposed method, we evaluated its performance with the five gold standard datasets presented in Table \ref{tb:SensEval-SemEval_Goldstandard}. We further present the results of the combined datasets to demonstrate the overall performance of the evaluated systems. The performance is measured by the F1-score discussed in Section \ref{subsec:ch4_Evaluation Metrics}. We present the proposed \gls{scsmm} method using two heuristics deployments; the first uses heuristics from the SemCor dataset ($H_s$), and the second uses both SemCor and \gls{omsti} datasets\footnote{The training dataset were downloaded from \url{http://lcl.uniroma1.it/wsdeval/training-data}} ($H_{so}$). In addition, we present three additional configurations for the \gls{scsmm} algorithm. These configurations demonstrate the effects of document context and document carry-forward on the performance of the proposed algorithm.
	
	Table \ref{tb:F1-score_DS} depicts the F1-score for each individual dataset in addition to the overall performance on all five datasets combined. The results of all configurations of the proposed \gls{scsmm} algorithm are compared to the current state-of-the-art knowledge-based systems presented in \cite{lesk1986WSD,agirre2009personalizing,agirre2014random,moro2014entity}. In addition, we present the baseline approaches using \textit{WN1\textsuperscript{st} sense}, \textit{\gls{mfs}\textsubscript{s}}, and \textit{\gls{mfs}\textsubscript{so}}.
	
	\begin{table}[!ht]
		\centering
		\caption{F1-score for each gold standard datasets}
		\label{tb:F1-score_DS}
		\begin{tabular}{p{.3\columnwidth}cccccc}
			\hline
			System		& SE2	& SE3 	& SE07 	& SE13 	& SE15 & All\\
			\hline				
			Lesk\textsubscript{ext}	&	50.6	&	44.5	&	32.0	&	53.6	&	51.0	&	48.7 \\
			Lesk\textsubscript{ext+emb}	&	63.0	&	63.7	&	56.7	&	66.2	&	64.6	&	63.7 \\
			UKB	&	56.0	&	51.7	&	39.0	&	53.6	&	55.2	&	53.2 \\
			UKB\textsubscript{gloss}	&	60.6	&	54.1	&	42.0	&	59.0	&	61.2	&	57.5 \\
			Babelfy	&	67.0	&	63.5	&	51.6	&	66.4	&	\textbf{70.3}	&	65.5 \\
			UKB\textsubscript{gloss18}	&	68.8	&	66.1	&	53.0	&	\textbf{68.8}	&	\textbf{70.3}	&	\textbf{67.3} \\
			WSD-TM	&	\textbf{69.0}	&	66.9	&	55.6	&	65.3	&	69.6	&	66.9 \\
			\hline
			WN1\textsuperscript{st} sense	&	66.8	&	66.2	&	55.2	&	63.0	&	67.8	&	65.2 \\
			MFS\textsubscript{s}	&	65.6	&	66	&	54.5	&	63.8	&	67.1	&	64.8  \\
			MFS\textsubscript{so}	&	66.5	&	60.4	&	52.3	&	62.6	&	64.2	&	62.8 \\
			\hline
			SCSMM\textsubscript{$H_{so}$}	&	66.9	&	67.2	&	55.4	&	63.0	&	68.4	&	65.6 \\
			SCSMM\textsubscript{$H_s$}	&	68.1	&	67.2	&	55.4	&	63.0	&	68.4	&	66.0 \\
			SCSMM\textsubscript{$H_s+DocCtx$}	&	68.4	&	66.8	&	56.9	&	63.4	&	69.0	&	66.2 \\
			SCSMM\textsubscript{$H_s+DocCF$}	&	68.1	&	67.1	&	56.3	&	63.0	&	68.7	&	66.0 \\
			SCSMM\textsubscript{$H_s+DocCtx+DocCF$}	&	68.9	&	\textbf{67.6}	&	\textbf{57.1}	&	63.5	&	69.5	&	66.7 \\				
			\hline
		\end{tabular}	
	\end{table}
	
	The proposed \gls{scsmm} algorithm has the best performance when the document context is included in the \gls{csm}, and when the \gls{doccf} disambiguation option is enabled. \gls{scsmm} outperforms all other systems on two datasets, the SE3 and SE07, while matching the WSD-TM system on SE2. 
	We noticed that our system is outperformed on SE13, as it is ranked fifth compared to other systems on the same datasets. We believe this is due to the following reasons: (1) This dataset is not diverse, as it includes only nouns, while with other datasets, various \gls{pos} contribute positively to the overall disambiguation algorithm. 
	However, we could not prove this causation due to the effects of other factors and the limited datasets. 
	(2) The other important factor is the average sentence size, as shown in Table \ref{tb:wsd_GD_Stats}. SE13 has an average sentence size of five terms per sentence, which is considered a small sentence size compared to other datasets. The only dataset that falls below that is SE07, which is explained next.
	
	Finally, the SE07 dataset has shown a consistent drop in performance across all systems. According to our analysis, this drop is due to three main reasons. Firstly, the high percentage of verbs within this dataset - because verbs have a very high granularity level, this has an inverse proportional effect on the disambiguation score (see Table \ref{tb:Goldstandard_Ambig_stats} and Fig. \ref{fig:Granularity_POS_Score}). Secondly, the dataset's small context size - the entire dataset contains two nouns and 24 verbs as a context, making the SE07 dataset the most ambiguous dataset with a 96\% ambiguity rate (see Table \ref{tb:wsd_GD_Stats}). Thirdly, and most importantly, the average sentence size - this dataset has the smallest average sentence size of three terms per sentence compared to all other datasets. Such a small average sentence size negatively impacts our algorithm because it identifies the global context between all terms, which is less accurate with shorter sentences.
	
	Additionally, Table \ref{tb:F1-score_POS} depicts the F1-score of the combined five datasets on each \gls{pos}. As can be seen from the results, our system outperforms all other systems when disambiguating nouns using the \gls{scsmm}\textsubscript{($H_s+DocCtx+DocCF$)} with a F1-score of 69.9. This is due to the proposed sequential algorithm that captures the maximum combination of the local similarities within each sentence. This can also be explained by the fact that nouns are structured and connected within WordNet compared to all other \gls{pos}. Note that \textit{Lesk\textsubscript{ext+emb} and WSD-TM} outperforms our system on verbs.
			
	\begin{table}[!ht]
		\centering
		\caption{F1-score for each POS on all gold standard datasets}
		\label{tb:F1-score_POS}
		\begin{tabular}{lcccc}
			\hline		
			System					&	Noun	&	Verb	&	Adj		&	Adv	\\
			\hline
			Lesk\textsubscript{ext}	&	54.1	&	27.9	&	54.6	&	60.3	 \\
			Lesk\textsubscript{ext+emb}	&	69.8	&	\textbf{51.2}	&	51.7	&	80.6	 \\
			UKB	&	56.7	&	39.3	&	63.9	&	44.0	 \\
			UKB\textsubscript{gloss}	&	62.1	&	38.3	&	66.8	&	66.2 \\
			Babelfy	&	68.6	&	49.9	&	73.2	&	79.8	 \\
			WSD-TM	&	69.7	&	\textbf{51.2}	&	\textbf{76.0}	&	\textbf{80.9}	 \\
			\hline
			WN1\textsuperscript{st} sense &	67.6	&	50.3	&	74.3	&	\textbf{80.9}	 \\
			MFS\textsubscript{s} 	&	67.6	&	49.6	&	73.1	&	80.5	 \\
			MFS\textsubscript{so}	&	65.8	&	45.9	&	72.7	&	80.5	 \\
			\hline
			SCSMM\textsubscript{$H_{so}$}	&	68.2	&	50.5	&	74.6	&	80.1	 \\
			SCSMM\textsubscript{$H_s$}	&	68.9	&	50.5	&	74.7	&	80.1	 \\
			SCSMM\textsubscript{$H_s+DocCtx$}	&	69.8	&	50.1	&	73.6	&	78.6	 \\
			SCSMM\textsubscript{$H_s+DocCF$}	&	68.9	&	50.8	&	74.5	&	80.1	 \\
			SCSMM\textsubscript{$H_s+DocCtx+DocCF$}	&	\textbf{69.9}	&	51.0	&	74.7	&	80.3	 \\				
			\hline
		\end{tabular}	
	\end{table}		
		
	\subsubsection{Discussion of Experimental Results}\label{subsubsec:Discussion of Experimental Results}
		Despite the various scores achieved by the evaluated systems, Table \ref{tb:F1-score_DS} shows a performance correlation across all systems. The results demonstrate a consensus on the best and worst scores per dataset. For instance, most systems perform best on \textit{SE15} and worst on \textit{SE07}. Based on the observation above, we present and analyze the effect of \gls{pos} distribution, granularity level, ambiguity rate, and sentence size on the performance of \gls{wsd} systems in general and the proposed \gls{scsmm} algorithm in particular. 	

\begin{figure}[!ht]
			\centering
			\includegraphics[trim=25 50 25 50,clip,width=.7\columnwidth]{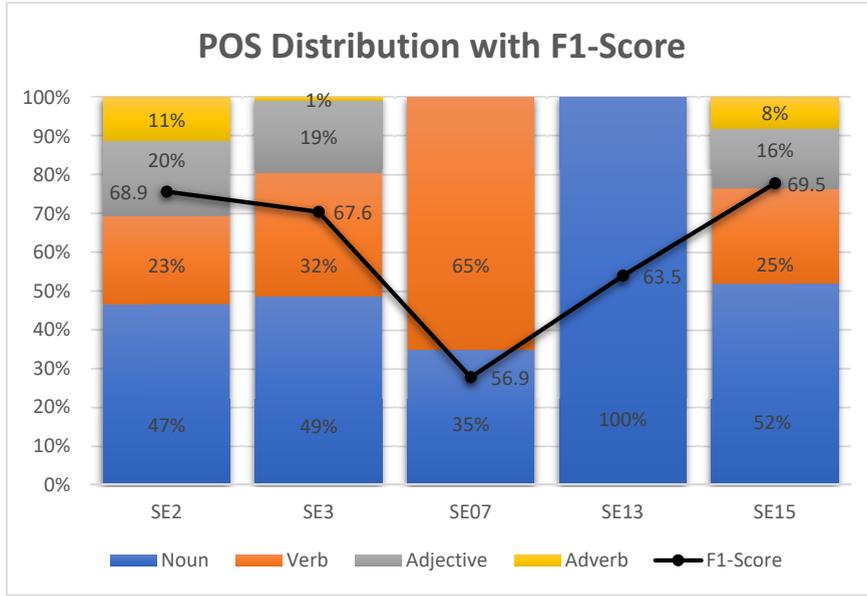}
			\caption{Distribution of POS compared to the F1-score}
			\label{fig:DS_POS_Distribution_Score}
		\end{figure}

		\textbf{POS Distribution}: The diversity of \gls{pos} within each dataset appears to correlate with the F1-score. Fig. \ref{fig:DS_POS_Distribution_Score} depicts the F1-score for our proposed \gls{scsmm} algorithm with the \gls{pos} distribution for each dataset. As shown in the figure, SE2 and SE15 contain similar \gls{pos} distribution, in particular, the weights of verbs within the datasets has a higher impact on the performance of any \gls{wsd} system, including the proposed algorithm. SE2 and SE15 contain almost the same percentage of verbs (23\% and 25\%), respectively, and have a similar F1-score. As for \textit{SE3}, verbs occupy 32\% of the dataset. Consequently, the performance of all systems has deteriorated for this dataset compared to \textit{SE2} and \textit{SE15}. Finally, having verbs outweigh nouns by almost double in \textit{SE07}, all systems showed the lowest F1-score on this dataset compared to all other datasets. 
		
\begin{figure}[!ht]
			\centering
			\includegraphics[trim=50 140 50 140,clip,width=.7\columnwidth]{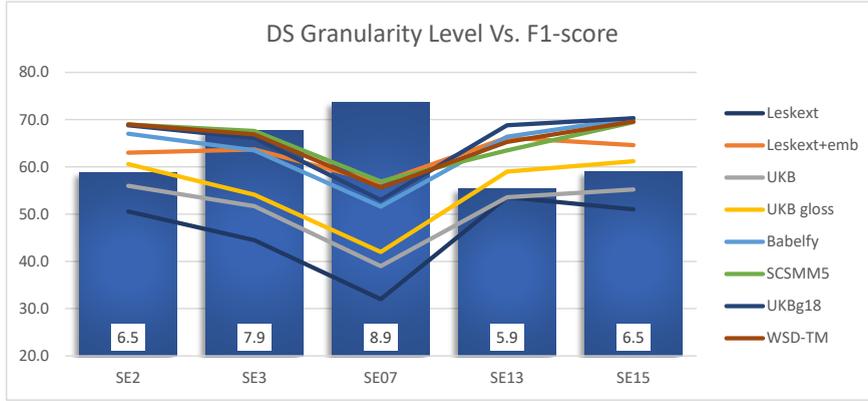}
			\caption{The granularity level of datasets compared to F1-score}
			\label{fig:Granularity_DS_Score}
		\end{figure}

		Finally, the trigger dataset for analyzing the \gls{pos} distribution is SE13. Although SE13 contains three of the best qualities a dataset could have, yet, it performs poorly compared to other diverse datasets. SE13 contains only nouns, which are well structured in WordNet. It has the lowest ambiguity rate (79\%, as shown in Table \ref{tb:wsd_GD_Stats}), and it has the lowest granularity level of 5.9 as a dataset (see Fig. \ref{fig:Granularity_DS_Score}). As a result, we conclude that a diverse distribution of \gls{pos} within a dataset improves our \gls{wsd} algorithm. 
				
\begin{figure}[!ht]
			\centering
			\includegraphics[trim=50 140 50 140,clip,width=.7\columnwidth]{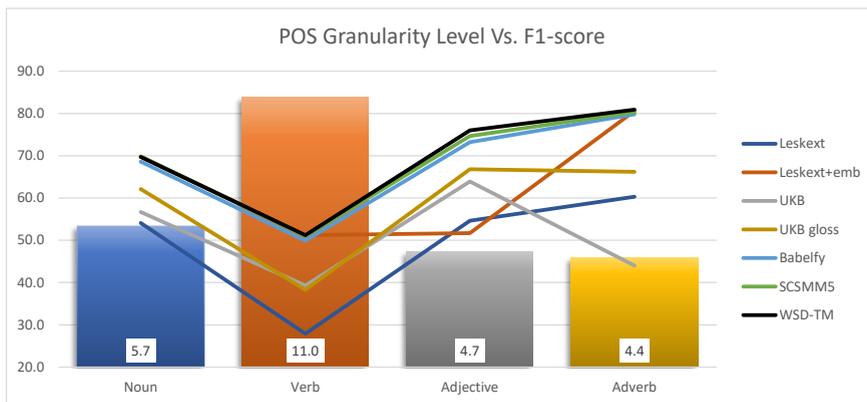}
			\caption{The granularity level of POS compared to F1-score}
			\label{fig:Granularity_POS_Score}
		\end{figure}

		\textbf{Granularity Level}: Granularity level is one of the most apparent factors that affect the performance of any \gls{wsd} system including the proposed algorithm. Fig. \ref{fig:Granularity_POS_Score} exhibits the performance of the proposed system and all other evaluated systems compared to the granularity level for each \gls{pos}. The columns in the figure represent the granularity levels, while the lines represent the F1-score for the evaluated systems. The figure clearly illustrates that the more granular senses within \gls{pos}, the lower the system's performance. 
		The same holds true for the granularity level within each dataset regardless of the \gls{pos} distribution. Fig. \ref{fig:Granularity_DS_Score} presents the F1-score for all systems on each dataset compared to the granularity level of each dataset.

\begin{figure}[!ht]
			\centering
			\includegraphics[trim=50 50 50 50,clip,width=.7\columnwidth]{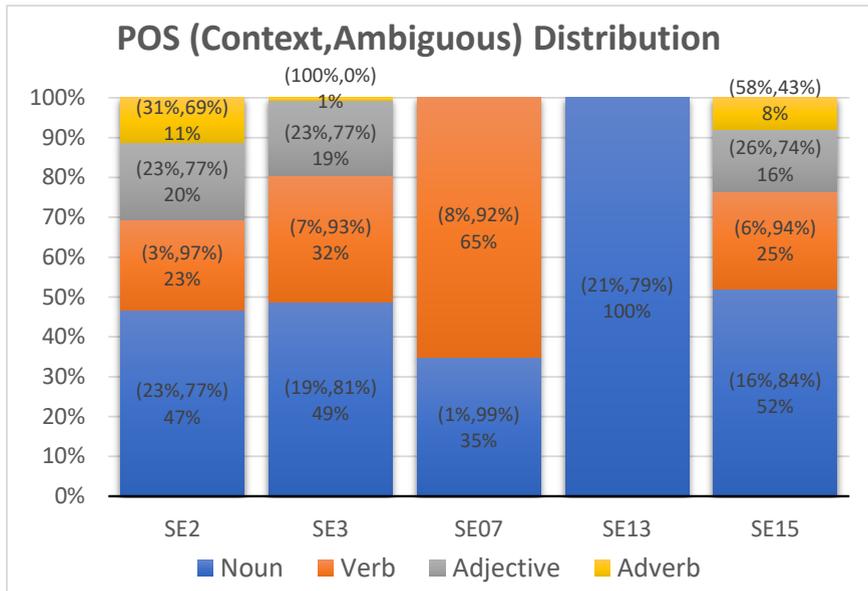}
			\caption{Distribution of POS with (context to ambiguous) ratio}
			\label{fig:DS_POS_CtxtoAmb_Distribution}
		\end{figure}

		\textbf{Context vs. Ambiguity Rates}: 	Both \textit{SE2} and \textit{SE15} have almost the same \gls{pos} distribution within their respective datasets (see Fig. \ref{fig:DS_POS_Distribution_Score}) and the exact same granularity level (see Fig. \ref{fig:Granularity_DS_Score}). On the other hand, the other three datasets have different \gls{pos} distribution and relatively higher granularity level. So what are the advantages of SE15 over SE2 that yield better performance? We believe this is due to the context and ambiguity rates. The ambiguity rate represents the percentage of ambiguous terms within each \gls{pos} or dataset. Fig. \ref{fig:DS_POS_CtxtoAmb_Distribution} depicts the \gls{pos} distribution for each dataset in addition to the context and ambiguity rates within each \gls{pos}. Except for the nouns, SE15 has a higher context rate than SE2, which explains the results of the F1-score for each \gls{pos} within these two datasets. Table \ref{tb:Se2_Se15_score} shows the F1-scores for the proposed \gls{scsmm}\textsubscript{($H_s+DocCtx+DocCF$)} algorithm for each \gls{pos} on SE2 and SE15 datasets. The results correlate with the context and ambiguity rates within each \gls{pos}. For example, SE2 has a higher context rate for the nouns than SE15; thus, it performed better. On the other hand, SE15 performed better than SE2 on all other \gls{pos}s due to their higher context rates. 
					
		\begin{table}[!ht]
			\centering
			\caption[F1-score for SCSMM per POS]{F1-score for SCSMM\textsubscript{($H_s+DocCtx+DocCF$)} per POS}
			\label{tb:Se2_Se15_score}
			\begin{tabular}{lcccc}
				\hline
				Dataset &Noun	&Verb	&Adjective	&Adverb	\\
				\hline
				SE2		&77.5	&43.3	&73.0	&78.0	\\
				SE15	&69.5	&57.4	&80.6	&85.0	\\		
				\hline
			\end{tabular}		
		\end{table}
		
		\textbf{Average Sentence Size}: 
		The average sentence size is the most important factor that affects the performance of our \gls{scsmm} algorithm and other systems, as it is more challenging to extract a context from fewer words. The same is true for a large number of words. The average sentence size is shown in Table \ref{tb:wsd_GD_Stats}, which explains the lower performance of SE13 compared to SE2, as the average sentence size is shorter for SE13. However, although SE15 has a shorter average sentence size than SE2, it performed better. This result can be justified by the context rate factor discussed above, or an indication of an optimal average sentence size. 

\section{Conclusion}\label{sec:Conclusion}
In this paper, we presented a novel knowledge-based \gls{wsd} approach. 
Unlike other systems, our proposed \gls{scsmm} algorithm exploits the merits of local context, word sense heuristics, and the global context while maintaining the words order. 
The proposed \gls{scsmm} algorithm exceeds the current state-of-the-art \gls{kg}-based systems when disambiguating nouns. Moreover, we evaluated the performance of current \gls{wsd} systems, including our proposed method, on well-known gold standard datasets from the SemEval workshop series. Based on the datasets analysis and the trends of the evaluated systems, we conclude that \gls{wsd} systems are negatively impacted by the granularity level of the dataset and the included \gls{pos}. On the other hand, a more diverse \gls{pos} within the dataset improves the results of the proposed \gls{wsd} algorithm. Similarly, the higher the context rate, the better the F1-score. Finally, the results show that very short sentences (i.e., fewer than three words) can negatively affect the proposed \gls{scsmm} algorithm.
	
We believe that as \gls{kg}s are enriched with more relationships between entities, and more domain-based \gls{kg} are exploited, knowledge-based systems will outperform other \gls{wsd} approaches. Furthermore, knowledge-based systems are intuitive, and their results are easily explained, understood, and justified by humans. 
The proposed method does not capture the exact topic of the document, but rather utilizes all context words in the document to disambiguate terms. To address this limitation, future research could investigate the adaptation of topic modeling and text clustering algorithms, such as the \gls{lda} algorithms used in \cite{chaplot2018knowledge}, or the $\beta$-hill climbing technique presented in \cite{abualigah2020improved} to improve the document context and its similarity with ambiguous terms. Future work could include the investigation of a comprehensive semantic similarity and relatedness measure, making use of both taxonomic and not-taxonomic relations existing in the \gls{kg} in order to capture true contextual relatedness between terms. 

\bibliographystyle{cas-model2-names}
\bibliography{mycollection}

\section*{Biography}
\bio[width=20mm,pos=l]{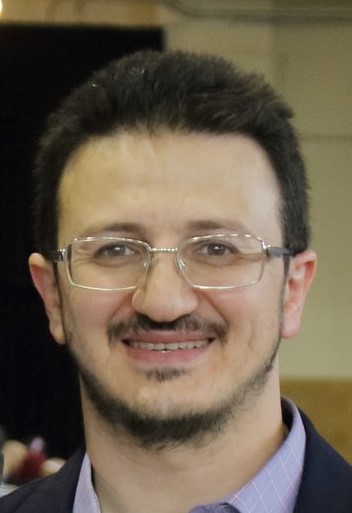}
\textbf{Mohannad AlMousa} received his Bachelor Degree from Ryerson University (Canada) in the field of Information Technology Management (specialized in Knowledge and database) in 2010. He then obtained a Master's Degree in Computer Science from Lakehead University (Canada) in 2014. Currently, he is a PhD candidate in Software Engineering at Lakehead University. Mr. AlMousa's current dissertation focuses on semantic similarity and relatedness and word sense disambiguation. His main research interests include Semantic knowledge representation, Recommender Systems, Natural Language Processing, and Knowledge Graphs.
\endbio
\vspace{.5cm}
\bio[width=20mm,pos=l]{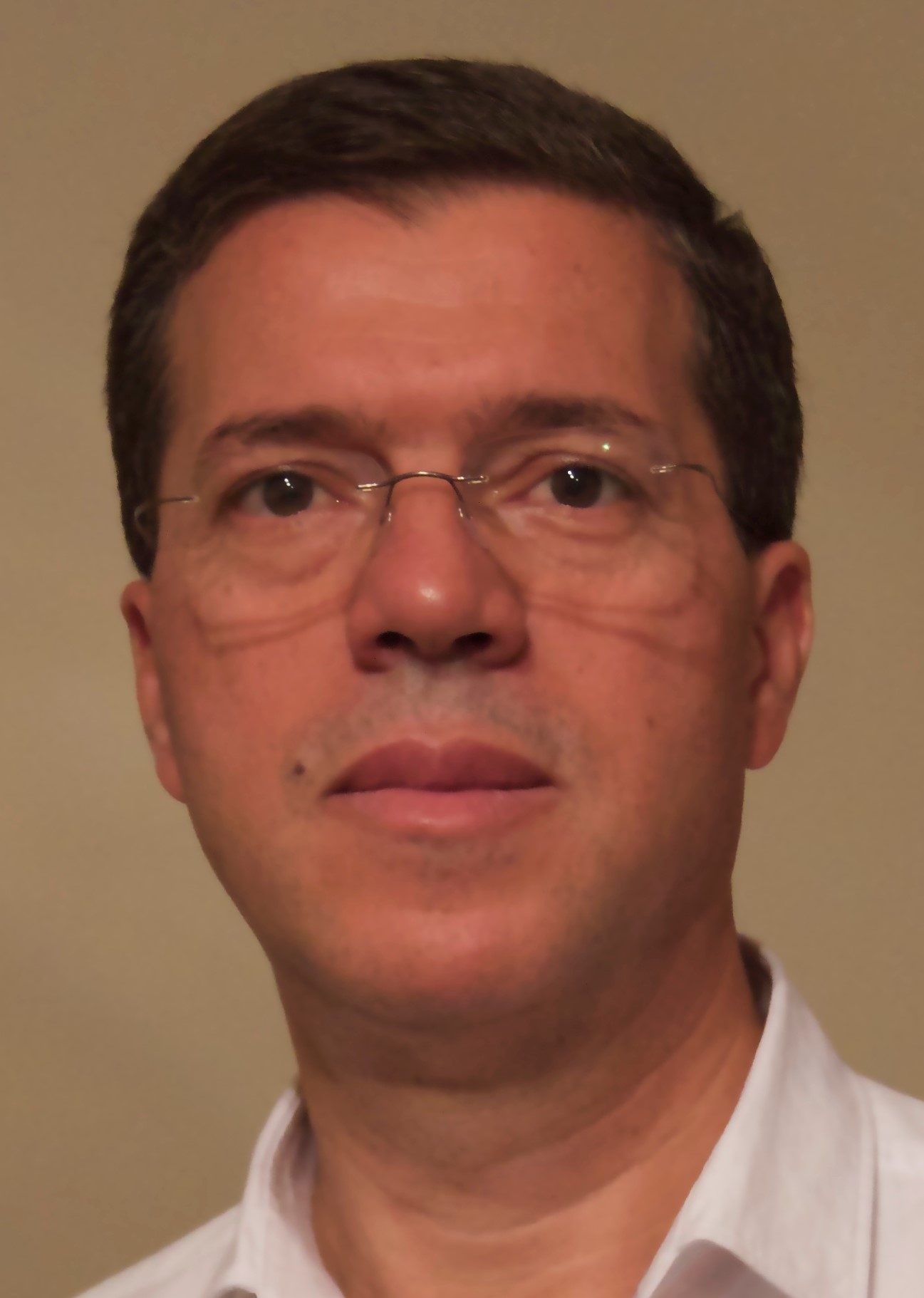} 
\textbf{Rachid Benlamri} is a Professor of Software Engineering at Lakehead University - Canada. He received his Master's degree and a PhD in Computer Science from the University of Manchester - UK in 1987 and 1990 respectively. He is the head of the Artificial Intelligence and Data Science Lab at Lakehead University. He supervised over 80 students and postdoctoral fellows. He served as keynote speaker and general chair for many international conferences. Professor Benlamri is a member of the editorial board for many referred international journals. His research interests are in the areas of Artificial Intelligence,Semantic Web,  Data Science, Ubiquitous Computing and Mobile Knowledge Management.
\endbio
\vspace{.5cm}
\bio[width=20mm,pos=l]{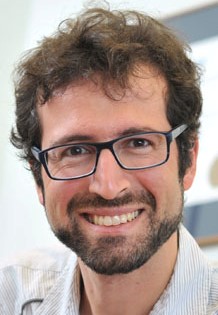}
\textbf{Richard Khoury} received his Bachelor’s Degree and his Master’s Degree in Electrical and Computer Engineering from Laval University (Québec City, QC) in 2002 and 2004 respectively, and his Doctorate in Electrical and Computer Engineering from the University of Waterloo (Waterloo, ON) in 2007. From 2008 to 2016, he worked as a faculty member in the Department of Software Engineering at Lakehead University. In 2016, he moved to Université Laval as an associate professor. Dr. Khoury’s primary areas of research are data mining and natural language processing, and additional interests include knowledge management, machine learning, and artificial intelligence.
\endbio
\end{document}

%% file: abrev.tex
\newacronym{lod}{LOD}{Linked Open Data}
\newacronym{kge}{KGE}{Knowledge Graph Embedding}
\newacronym{nlp}{NLP}{Natural Language Processing}
\newacronym{nltk}{NLTK}{Natural Language Toolkit}
\newacronym{mt}{MT}{Machine Translation}
\newacronym{ic}{IC}{Information Content}
\newacronym{wsd}{WSD}{Word Sense Disambiguation}
\newacronym{wsi}{WSI}{Word Sense Induction}
\newacronym{kb}{KB}{knowledge base}
\newacronym{sg}{SG}{semantic graph}
\newacronym{kg}{KG}{Knowledge Graph}
\newacronym{isa}{ISA}{IS A}
\newacronym{ir}{IR}{Information Retrieval}
\newacronym{pr-ssr}{PR-SSR}{Poly-Relational Semantic Similarity and Relatedness}
\newacronym{pos}{POS}{Part Of Speech}
\newacronym{ric}{RIC}{Relation Information Content}
\newacronym{semic}{SemIC}{Semantic Information Content}
\newacronym{rdf}{RDF}{Resource Description Framework}
\newacronym{lcs}{LCS}{Least Common Subsumer}
\newacronym{ner}{NER}{Named Entity Recognition}
\newacronym{qa}{QA}{Question Answering}
\newacronym{ml}{ML}{Machine Learning}
\newacronym{dl}{DL}{Decision List}
\newacronym{dt}{DT}{Decision Trees}
\newacronym{nb}{NB}{Naive Bayes}
\newacronym{svm}{SVM}{Support Vector Machine}
\newacronym{nn}{NN}{Neural Network}
\newacronym{ims}{IMS}{``it makes sense"}
\newacronym{lstm}{LSTM}{Long Short-Term Memory}
\newacronym{bilstm}{BiLSTM}{Bidirectional Long Short-Term Memory}
\newacronym{cbc}{CBC}{Clustering by Committee}
\newacronym{omsti}{OMSTI}{One Million Sense-Tagged Instances)}
\newacronym{dfs}{DFS}{Depth First Search}
\newacronym{scsmm}{SCSMM}{Sequential Contextual Similarity Matrix Multiplication}
\newacronym{csm}{CSM}{Contextual Similarity Matrix}
\newacronym{ls}{LS}{Lexical Sample}
\newacronym{aw}{AW}{All-Words}
\newacronym{mfs}{MFS}{Most Frequent Sense}
\newacronym{doccf}{DocCF}{Document Carry Forward}
\newacronym{tfidf}{TF-IDF}{Term Frequency-Inverse Document Frequency}
\newacronym{idf}{IDF}{Inverse Document Frequency}
\newacronym{lsa}{LSA}{Latent Semantic Analysis}
\newacronym{mse}{MSE}{Mean-Squared Error}
\newacronym{ssd}{SSD}{Summation of Squared Difference}
\newacronym{igf}{IGF}{Inverse Glass Frequency}
\newacronym{lda}{LDA}{Latent Dirichlet Allocation}
\makeglossaries